\documentclass[runningheads]{llncs}

\usepackage{eccv}

\usepackage{eccvabbrv}

\usepackage{graphicx}
\usepackage{booktabs}

\usepackage{placeins}
\usepackage{float}
\usepackage{colortbl} 
\usepackage{xspace}
\usepackage[dvipsnames]{xcolor}
\usepackage{diagbox} 
\usepackage{multirow}
\usepackage{hhline}
\usepackage{enumitem}

\usepackage[accsupp]{axessibility}  % Improves PDF readability for those with disabilities.

\usepackage{hyperref}

\usepackage{orcidlink}

\makeatletter
\newcommand{\printfnsymbol}[1]{%
        \textsuperscript{\@fnsymbol{#1}}%
}
\makeatother

\begin{document}

% ---------------------------------------------------------------
\title{Per-Gaussian Embedding-Based Deformation for Deformable 3D Gaussian Splatting} 
\titlerunning{E-D3DGS}

\author{Jeongmin Bae\inst{1}\thanks{Authors contributed equally to this work.}\orcidlink{0009-0009-3376-2275}% 
\and
Seoha Kim\inst{1}\printfnsymbol{1}\orcidlink{0009-0006-7456-701X} 
\and
Youngsik Yun\inst{1}\orcidlink{0000-0003-4398-7856}
\and \\
Hahyun Lee\inst{2}\orcidlink{0000-0001-7043-7564}
\and
Gun Bang\inst{2}\orcidlink{0000-0003-4355-599X}
\and
Youngjung Uh\inst{1}\orcidlink{0000-0001-8173-3334}
}

\authorrunning{J. Bae et al.}

\institute{Yonsei University, Seoul 03722, Korea \\
\email{\{jaymin.bae, hailey07, bbangsik, yj.uh\}@yonsei.ac.kr} \and
Electronics and Telecommunications Research Institute, Daejeon 34129, Korea \\
\email{\{hanilee, gbang\}@etri.re.kr}
}

\newcommand{\needcite}[1]{\textcolor{red}{[CITE #1]}}
\newcommand{\temp}[1]{\textcolor[rgb]{0.032, 0.6392, 0.2039}{{\textbf{#1}}}}

\newcommand{\sh}[1]{\textcolor{violet}{#1}}
\newcommand{\shc}[1]{\textcolor{violet}{[sh: #1]}}

\newcommand{\bb}[1]{\textcolor{Bittersweet}{#1}}
\newcommand{\bbc}[1]{\textcolor{Bittersweet}{\tiny{[bb: #1]}}}

\newcommand{\uh}[1]{\textcolor{teal}{#1}} 
\newcommand{\uhc}[1]{\textcolor{red}{\tiny{[uh: #1]}}}

\newcommand{\jm}[1]{\textcolor{blue}{#1}} 
\newcommand{\jmc}[1]{\textcolor{blue}{[jm: #1]}}

\newcommand{\Fref}[1]{Figure \ref{#1}}
\newcommand{\fref}[1]{Figure \ref{#1}}
\newcommand{\Sref}[1]{Section \ref{#1}}
\newcommand{\sref}[1]{Section \ref{#1}}  % {$\S$\ref{#1}}

\newcommand{\Tref}[1]{Table \ref{#1}}
\newcommand{\tref}[1]{Table \ref{#1}}
\newcommand{\Aref}[1]{Appendix \ref{#1}}
\newcommand{\aref}[1]{Appendix \ref{#1}}
\newcommand{\Eref}[1]{Eq. \ref{#1}}
\newcommand{\eref}[1]{Eq. \ref{#1}}
\newcommand{\todo}[1]{\textbf{\textcolor{red}{[todo: #1]}}}
\newcommand{\xmark}{\ding{53}}
\newcommand{\degree}{\ensuremath{^\circ}}

\newcommand{\mypara}[1]{\noindent\textbf{#1}}

\newcommand{\myparagraph}[1]{\subsubsection{#1}}
\newcommand{\myfirstparagraph}[1]{\noindent{\textbf{#1} {} {}}}

\newcommand{\lorem}{\temp{Lorem ipsum dolor sit amet, consectetur adipiscing elit, sed do eiusmod tempor incididunt ut labore et dolore magna aliqua.}}
\newcommand{\shortlorem}{\temp{Neque porro quisquam est qui dolorem ipsum quia dolor sit amet, consectetur, adipisci velit.}}

\newcommand{\best}{\cellcolor[rgb]{0.96,0.8,0.8}} 
\newcommand{\second}{\cellcolor[rgb]{0.99,0.9,0.8}} 

% ours
\newcommand{\ours}{E4DGS\xspace}
\newcommand{\rot}{\mathbf{r}}
\newcommand{\scale}{\mathbf{s}}

\newcommand{\deformc}{\mathcal{F}_{\theta_\text{c}}}
\newcommand{\deformf}{\mathcal{F}_{\theta_\text{f}}}
\newcommand{\embg}{\mathbf{z}_\text{g}}
\newcommand{\embt}{\mathbf{z}_t}
\newcommand{\embtc}{\mathbf{z}_t^{_\text{c}}}
\newcommand{\embtf}{\mathbf{z}_t^{_\text{f}}}

% notation - nerf
\newcommand{\dir}{\mathbf{d}}  % viewing direction
\newcommand{\pos}{\mathbf{x}}  % 3d coordinates
\newcommand{\pe}{\gamma{}}  % positional encoding function
\newcommand{\ray}{\mathbf{r}}  % viewing direction
\newcommand{\C}{\mathbf{C}}  % gt rgb pixel
\newcommand{\Chat}{\hat{\mathbf{C}}}  % rendered rgb pixel
%  rendering equation
\newcommand{\mlpparam}{\Theta}  % paramset of field mlp
\newcommand{\T}{T}  % accumulated transmittance
\newcommand{\density}{\sigma}  % 
\newcommand{\radiance}{\mathbf{c}}  % 
\newcommand{\interval}{\delta}  % distance btw two adjacent points along the ray

\newcommand{\ourtime}{t'_{c}}  % ray
\newcommand{\offsets}{\boldsymbol{\delta}}  % ray
\newcommand{\offset}{\delta}  % ray

\newcommand{\z}{\mathbf{z}}  % random noise
\newcommand{\f}{\mathbf{f}}  % feature vector at 3D position
\newcommand{\Fhat}{\hat{\mathbf{F}}}  % rendered feature pixel

\newcommand{\E}{E}  % patch encoder
\newcommand{\D}{D}  % rgb decoder
\newcommand{\eps}{\epsilon}  % noise
\newcommand{\normal}{\mathcal{N}}  % noise
\newcommand{\patch}{\mathbf{I}_{\text{p}}}  % rgb decoder

% losses
\newcommand{\Lact}{\mathcal{L}_{\text{act}}}  % activation regularization
\newcommand{\Lrecon}{\mathcal{L}_{\text{recon}}}  % rgb recon loss
\newcommand{\Lrender}{\mathcal{L}_{\text{render}}}  % nerf rendering loss

% math
\newcommand{\real}{\mathbb{R}}

% added by jm
\newcommand{\cmark}{\ding{51}}%

\maketitle
\begin{figure}[h!]
\centering
\includegraphics[width=1\linewidth]{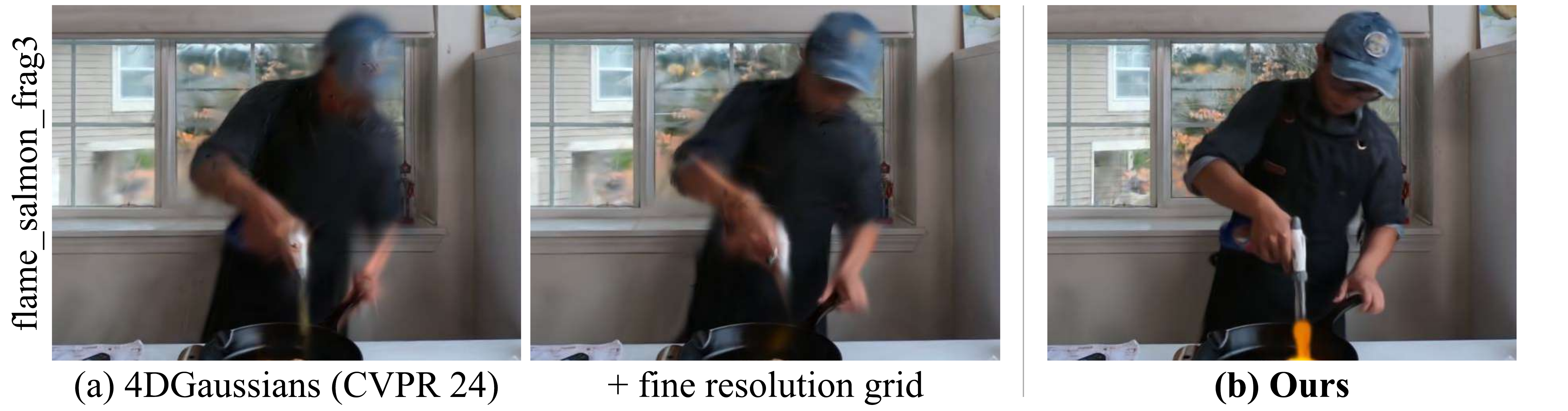}
\caption{\textbf{Overview.} (a) Existing deformable 3D Gaussian Splatting methods show blurry results in complex dynamic scenes, even with deformation fields using finer feature grids. (b) Our model solves the problem by employing per-Gaussian latent embeddings to predict deformations for each Gaussian and achieves clearer results.}
\label{fig:teaser}
\end{figure}
\begin{abstract}
As 3D Gaussian Splatting (3DGS) provides fast and high-quality novel view synthesis, it is a natural extension to deform a canonical 3DGS to multiple frames for representing a dynamic scene. However, previous works fail to accurately reconstruct complex dynamic scenes. We attribute the failure to the design of the deformation field, which is built as a coordinate-based function. This approach is problematic because 3DGS is a mixture of multiple fields centered at the Gaussians, not just a single coordinate-based framework. To resolve this problem, we define the deformation as a function of per-Gaussian embeddings and temporal embeddings. Moreover, we decompose deformations as coarse and fine deformations to model slow and fast movements, respectively. Also, we introduce a local smoothness regularization for per-Gaussian embedding to improve the details in dynamic regions. 

\: Project page: \href{https://jeongminb.github.io/e-d3dgs/}{https://jeongminb.github.io/e-d3dgs/}


\end{abstract}

%%%%%%%%%%%%%%%%%%%%%%%%%%%%%%%%%%%%%%%%%%%%%%%%
%%%%%%%%%%%%%%%%%%%%%%%%%%%%%%%%%%%%%%%%%%%%%%%%
%%%%%%%%%%%%%%%%%%%%%%%%%%%%%%%%%%%%%%%%%%%%%%%%

\section{Introduction}
\label{sec:intro}
Dynamic scene reconstruction from multi-view input videos is an important task in computer vision, as it can be extended to various applications and industries such as mixed reality, content production, \textit{etc}. Neural Radiance Fields (NeRF) \cite{mildenhall2021nerf}, which enable photorealistic novel view synthesis from multi-view inputs, can represent dynamic scenes by modeling the scene with an additional time input \cite{pumarola2020d,li2022neural}. However, typical NeRFs require querying multilayer perceptron (MLP) for hundreds of points per camera ray, which limits rendering speed.

On the other hand, the recently emerging 3D Gaussian Splatting (3DGS) \cite{kerbl3Dgaussians} has the advantage of real-time rendering compared to NeRFs using a differentiable rasterizer for 3D Gaussian primitives. 3DGS directly optimizes the parameters of 3D Gaussians (position, opacity, anisotropic covariance, and spherical harmonics coefficients) and renders them via projection and $\alpha$-blending. Since 3DGS has the characteristics of continuous volumetric radiance fields, some recent studies\cite{wu20234dgaussians,yang2023deformable3dgs,duisterhof2023md,huang2023sc,liang2023gaufre,yu2023cogs} represent dynamic scenes by defining a canonical 3DGS and deforming it to individual frames as deformable NeRFs \cite{yang2023deformable3dgs} do. Specifically, they model the deformation as a function of 4D (x, y, z, t) coordinates with MLPs or grids to predict the change in the 3D Gaussian parameters. 

However, since 3DGS is a mixture of multiple volumetric fields, it is not appropriate to model the deformation of Gaussian parameters with a single coordinate-based network to represent dynamic scenes. In addition, existing field-based approaches are constrained by the resolution of the grid which models the deformation field, the capacity of the model, or the frequencies of the input. As shown in \fref{fig:teaser}, the existing study does not properly represent complex dynamic scenes, and even introducing an additional feature grid that is twice the maximum resolution has only a slight improvement in performance (See Appendix for more results). We alleviate this problem by introducing a novel dynamic representation to deform each Gaussian.

In this paper, we model the deformation of Gaussians at frames as 1) a function of a product space of per-Gaussian embeddings and temporal embeddings. We expect this rational design to bring quality improvement by precisely modeling different deformations of different Gaussians. Additionally, 2) We decompose temporal variations of the parameters into coarse and fine components, namely coarse-fine deformation. The coarse deformation represents large or slow movements in the scene, while fine deformation learns the fast or detailed movements that coarse deformation does not cover. Finally, we propose 3) a local smoothness regularization for per-Gaussian embedding to ensure the deformations of neighboring Gaussians are similar.

In our experiments, we observe that our per-Gaussian embeddings, coarse-fine deformation, and regularization improve the deformation quality. Our approach outperforms baselines in capturing fine details in dynamic regions and excels even under challenging camera settings. Additionally, our method also achieves fast rendering speed and relatively low capacity.

%%%%%%%%%%%%%%%%%%%%%%%%%%%%%%%%%%%%%%%%%%%%%%%%
%%%%%%%%%%%%%%%%%%%%%%%%%%%%%%%%%%%%%%%%%%%%%%%%
%%%%%%%%%%%%%%%%%%%%%%%%%%%%%%%%%%%%%%%%%%%%%%%%

\section{Related Work}
In this section, we review methods for dynamic scene reconstruction that deform 3D canonical space and methods for reconstructing dynamic scenes utilizing dynamic 3D Gaussians. Afterward, we review methods that use embeddings and spatial relationships of Gaussians.

%%%%%%%%%%%%%%%%%%%%%%%%%%%%%%%%%%%%%%%%%%%%%%%%

\myparagraph{Deforming 3D Canonical Space}
D-NeRF\cite{pumarola2020d} reconstructs dynamic scenes by deforming ray samples over time, using the deformation network that takes 3D coordinates and timestamps of the sample as inputs. Nerfies\cite{park2021nerfies} and HyperNeRF\cite{park2021hypernerf} use per-frame trainable deformation codes instead of time conditions to deform the canonical space. Instead of deforming from the canonical frame to the entire frames, HyperReel\cite{attal2023hyperreel} deforms the ray sample of the keyframe to represent the intermediate frame. 4DGaussians \cite{wu20234dgaussians} and D3DGS \cite{yang2023deformable3dgs} reconstruct the dynamic scene with a deformation network which inputs the center position of the canonical 3D Gaussians and timestamps. MoDGS\cite{liu2024modgs} learns the mapping between canonical space and space at a specific timestamp through invertible MLP. In contrast, we demonstrate a novel deformation representation as a function of a product space of per-Gaussian latent embeddings and temporal embeddings.

%%%%%%%%%%%%%%%%%%%%%%%%%%%%%%%%%%%%%%%%%%%%%%%%

\myparagraph{Dynamic 3D Gaussians}
To extend the fast rendering speed of 3D Gaussian Splatting\cite{kerbl3Dgaussians} into dynamic scene reconstructions. 4DGaussians\cite{wu20234dgaussians} decodes features from multi-resolution HexPlanes \cite{cao2023hexplane} for temporal deformation of 3D Gaussians. While D3DGS\cite{yang2023deformable3dgs} uses an implicit function that processes the time and location of the Gaussian. 4DGS\cite{yang2023gs4d} decomposes the 4D Gaussians into a time-conditioned 3D Gaussians and a marginal 1D Gaussians. STG\cite{li2023spacetimegaussians} represents changes in 3D Gaussian over time through a temporal opacity and a polynomial function for each Gaussian.

Our method uses deformable 3D Gaussians as  4DGaussians\cite{wu20234dgaussians} and D3DGS\cite{yang2023deformable3dgs} do, but does not necessitate the separated feature field to obtain the input feature of the deformation decoder. Our approach uses embeddings allocated to each Gaussian and a temporal embedding shared within a specific frame.

%%%%%%%%%%%%%%%%%%%%%%%%%%%%%%%%%%%%%%%%%%%%%%%%

\myparagraph{Latent Embedding on Novel View Synthesis}
Some studies incorporate latent embeddings to represent different states of the static and dynamic scene. NeRF-W\cite{martinbrualla2020nerfw} and Block-NeRF\cite{tancik2022blocknerf} employ per-image embeddings to capture different appearances of a scene, representing the scenes from in-the-wild image collections. DyNeRF and MixVoxels\cite{li2022neural,wang2022mixed} employ a temporal embedding for each frame to represent dynamic scenes. Nerfies\cite{park2021nerfies} and HyperNeRF\cite{park2021hypernerf} incorporate both per-frame appearance and deformation embeddings. Sync-NeRF\cite{kim2023sync} introduces time offset to calibrate the misaligned temporal embeddings on dynamic scenes from unsynchronized videos. We introduce per-Gaussian latent embedding to encode the changes over time of each Gaussian and use temporal embeddings to represent different states in each frame of the scene. 

%%%%%%%%%%%%%%%%%%%%%%%%%%%%%%%%%%%%%%%%%%%%%%%%

\myparagraph{Considering Spatial Relationships of Gaussians}
Scaffold-GS \cite{scaffoldgs} reconstructs 3D scenes by synthesizing Gaussians from anchors, utilizing that the neighboring Gaussians have similar properties. SAGS \cite{ververas2024sags} creates a graph based on k-nearest neighbors (KNN) so that each Gaussian is optimized while considering its neighboring Gaussians. In dynamic scene reconstruction, SC-GS \cite{Huang_2024_CVPR} and GaussianPrediction \cite{Zhao_2024} deform Gaussians by combining the deformations of key point Gaussians. Dynamic 3D Gaussians \cite{luiten2023dynamic} utilizes regularization to encourage that Gaussians and their neighboring Gaussians deform with local rigidity. Similarly, we propose a local smoothness regularization that encourages neighboring Gaussians to have similar embeddings, resulting in similar deformations.

%%%%%%%%%%%%%%%%%%%%%%%%%%%%%%%%%%%%%%%%%%%%%%%%
%%%%%%%%%%%%%%%%%%%%%%%%%%%%%%%%%%%%%%%%%%%%%%%%
%%%%%%%%%%%%%%%%%%%%%%%%%%%%%%%%%%%%%%%%%%%%%%%%

\section{Method}

In this section, we first provide a brief overview of 3D Gaussian Splatting (\sref{sec:prelim}). 
Next, we introduce our overall framework, embedding-based deformation for Gaussians (\sref{sec:embedding}) and coarse-fine deformation scheme consisting of coarse and fine deformation functions (\sref{sec:deformation}). Finally, we present a local smoothness regularization for per-Gaussian embeddings to achieve better details on dynamic regions (\sref{sec:reg}). 
%we present an efficient training strategy for faster convergence and higher performance (\sref{sec:training}).

%%%%%%%%%%%%%%%%%%%%%%%%%%%%%%%%%%%%%%%%%%%%%%%%

\subsection{Preliminary: 3D Gaussian Splatting}
\label{sec:prelim}
3D Gaussian splatting\cite{kerbl3Dgaussians} optimizes a set of anisotropic 3D Gaussians through differentiable tile rasterization to reconstruct a static 3D scene. By its efficient rasterization, the optimized model enables real-time rendering of high-quality images.
Each 3D Gaussian kernel $G_i(x)$ at the point $x$ consist with position $\pos_i$, rotation $R_i$, and scale $S_i$:
\begin{equation}
    G_i(x) = e^{-\frac{1}{2}(x-\pos_i)^T\Sigma_i^{-1}(x-\pos_i)},  \qquad \text{where} \quad \Sigma_i =  R_iS_iS_i^TR_i^T.
    \label{eq:gaussian}
\end{equation}
To projecting 3D Gaussians to 2D for rendering, covariance matrix $\Sigma'$ are calculated by viewing transform $W$ and the Jacobian $J$ of the affine approximation of the projective transfomation\cite{zwicker2001surface} as follows:
\begin{equation}
    \Sigma' = J W \Sigma W^{T} J^{T}.
    \label{eq:cov2d}
\end{equation}
Blending $\mathcal{N}$ depth-ordered projected points that overlap the pixel, the Gaussian kernel $G_i(x)$ is multiplied by the opacity of the Gaussian $\sigma_i$ and calculates the pixel color $C$ with the color of the Gaussian $c_i$:
\begin{equation}
    C = \sum_{i \in \mathcal{N}} c_i \alpha_i \prod_{j=1}^{i-1} (1 - \alpha_j), \qquad \text{where} \quad \alpha_i = \sigma_iG_i(x).
    \label{eq:blending}
\end{equation}
The color of Gaussian $c_i$ is determined using the SH coefficient with the viewing direction.

%%%%%%%%%%%%%%%%%%%%%%%%%%%%%%%%%%%%%%%%%%%%%%%%

\begin{figure}[t]
\centering
\includegraphics[width=1\linewidth]{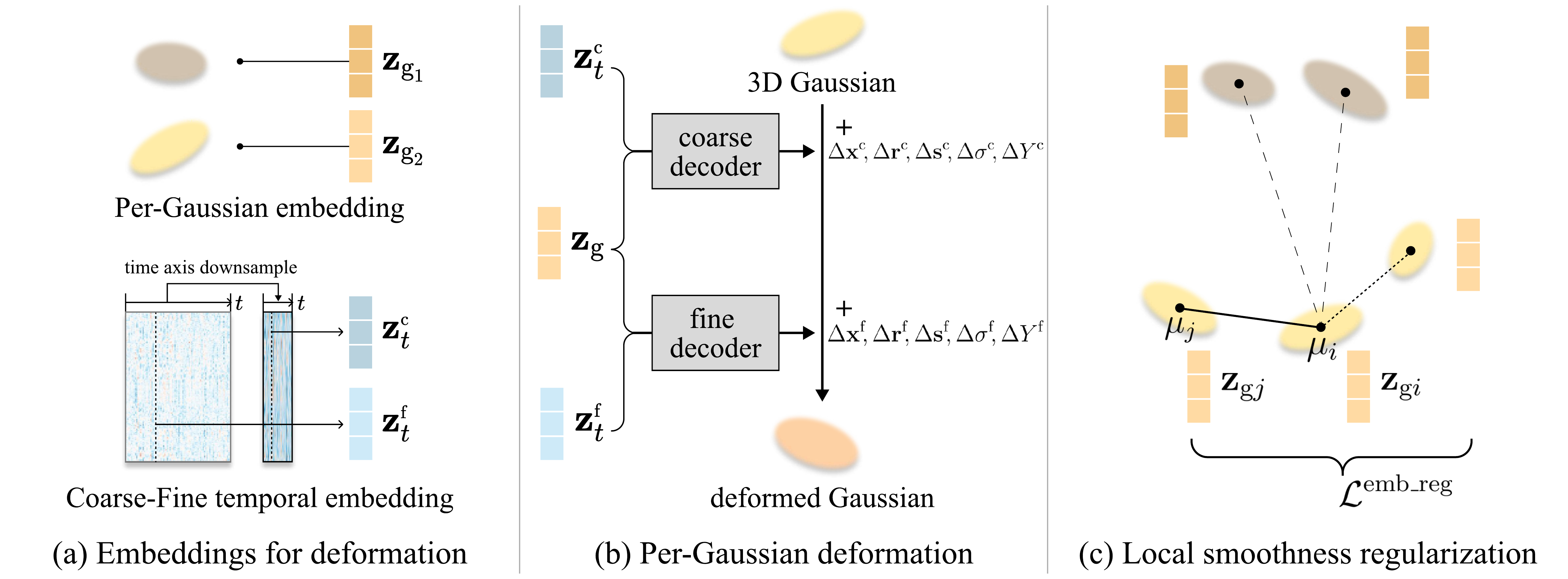}
\caption{\textbf{Framework.} Existing coordinate-based network methods struggle to represent complex dynamic scenes. To this end, we define per-Gaussian deformation. (a) Firstly, we assign a latent embedding for each Gaussian. Additionally, we introduce coarse and fine temporal embeddings to represent the slow and fast state of the dynamic scene. (b) By employing two decoders that take per-Gaussian latent embeddings along with coarse and fine temporal embeddings as input, we estimate slow or large changes and fast or detailed changes to model the final deformation, respectively. (c) Finally, we introduce a local smoothness regularization so that the embeddings of neighboring Gaussians are similar.}
\label{fig:framework}
\end{figure}

\subsection{Embedding-Based Deformation for Gaussians}
\label{sec:embedding}

Deformable NeRFs consist of a deformation field that predicts displacement $\Delta \pos$ for a given coordinate $\pos$ from the canonical space to each target frame, and a radiance field that maps color and density from a given coordinate in the canonical space ($\pos + \Delta \pos$). 
Existing deformable Gaussian methods employ the same approach for predicting the deformation of Gaussians, i.e., utilizing a deformation field based on coordinates.

Unlike previous methods, we start from the design of 3DGS: the 3D scene is represented as a mixture of Gaussians that have individual radiance fields. Accordingly, the deformation should be defined for each Gaussian. Based on this intuition, we introduce a function $\mathcal{F}_\theta$ that produces deformation from learnable embeddings $\embg\in\real^{32}$ belonging to individual Gaussians (\fref{fig:framework}a), and typical temporal embeddings $\embt\in\real^{256}$ for different frames:
\begin{equation}
    \mathcal{F}_\theta: (\embg, \embt) \rightarrow (\Delta \pos, \Delta \rot, \Delta \scale, \Delta \sigma, \Delta Y),
    \label{eq:deform_our}
\end{equation}
where $\rot$ is a rotation quaternion, $\scale$ is a vector for scaling, $\sigma$ is an opacity, and $Y$ is SH coefficients for modeling view-dependent color.
We implement $\mathcal{F}_\theta$ as a shallow multi-layer perceptron (MLP) followed by an MLP head for each parameter. As a result, the Gaussian parameters at frame $t$ are determined by adding $\mathcal{F}_\theta(\embg, \embt)$ to the canonical Gaussian parameters (\fref{fig:framework}c).

We jointly optimize the per-Gaussian embeddings $\embg$, the deformation function $\mathcal{F}_\theta$, and the canonical Gaussian parameters to minimize the rendering loss. We use the L1 and periodic DSSIM as the rendering loss between the rendered image and the ground truth image.

%%%%%%%%%%%%%%%%%%%%%%%%%%%%%%%%%%%%%%%%%%%%%%%%

\subsection{Coarse-Fine Deformation}
\label{sec:deformation}

Different parts of a scene may have coarse and fine motions \cite{feichtenhofer2019slowfast}. E.g., a hand swiftly stirs a pan (fine) while a body slowly moves from left to right (coarse). Based on this intuition, we introduce a coarse-fine deformation that produces a summation of coarse and fine deformations.

Coarse-fine deformation consists of two functions with the same architecture: one for coarse and one for fine deformation (\fref{fig:framework}c). The functions receive different temporal embeddings as follows:

Following typical temporal embeddings, we start from a 1D feature grid $Z\in\real^{N \times 256}$ for $N$ frames and use an embedding $\embtf=\text{interp}(Z,t)$ for fine deformation. For coarse deformation, we linearly downsample $Z$ by a factor of 5 to remove high-frequencies responsible for fast and detailed deformation. Then we compute $\embtc$ as a linear interpolation of embeddings at enclosing grid points (\fref{fig:framework}b).

As a result, coarse deformation $\deformc$($\embg$, $\embtc$) is responsible for representing large or slow movements in the scene, while fine deformation $\deformf$($\embg$, $\embtf$) learns the fast or detailed movements that coarse deformation does not cover. This improves the deformation quality. Refer to the Ablation study section for more details.

%%%%%%%%%%%%%%%%%%%%%%%%%%%%%%%%%%%%%%%%%%%%%%%%

\subsection{Local Smoothness Regularization}
\label{sec:reg}

Neighboring Gaussians constructing dynamic objects tend to exhibit locally similar deformation.
Inspired by \cite{luiten2023dynamic}, we introduce a local smoothness regularization for per-Gaussian embedding $\embg$ (\fref{fig:framework}d) to encourage similar deformations between nearby Gaussians $i$ and $j$:
$$\mathcal{L}^{\text{emb\_reg}}= \frac{1}{k|\mathcal{S}|} \sum_{i\in \mathcal{S}} \sum_{j\in \text{KNN}_{i;k}} (w_{i,j}\| \mathbf{z}_\text{g}{}_i - \mathbf{z}_\text{g}{}_j \|_2),$$ 
\text{where} $w_{i,j}=\exp(-\lambda_{w} \| \mu_{j} - \mu_{i} \|^2_2)$ is the weighting factor and $\mu$ is the Gaussian center. We set $\lambda_{w}$ to 2000 and $k$ to 20 following \cite{luiten2023dynamic}. To reduce the computational cost, we obtain sets of k-nearest-neighbors only when the densification occurs.

Note that unlike previous approaches that directly constrain physical properties such as rigidity or rotation, We implicitly induce locally similar deformation by ensuring that per-Gaussian embeddings are locally smooth. Our regularization allows better capture of textures and details of dynamic objects.

%%%%%%%%%%%%%%%%%%%%%%%%%%%%%%%%%%%%%%%%%%%%%%%%
%%%%%%%%%%%%%%%%%%%%%%%%%%%%%%%%%%%%%%%%%%%%%%%%
%%%%%%%%%%%%%%%%%%%%%%%%%%%%%%%%%%%%%%%%%%%%%%%%

\section{Experiment}

In this section, we first describe the criterion for selection of baselines, and evaluation metrics. We then demonstrate the effectiveness of our method through comparisons with various baselines and datasets (\sref{sec:dynerf}-\ref{sec:hypernerf}). Finally, we conduct analysis and ablations of our method (\sref{sec:analysis}).

%%%%%%%%%%%%%%%%%%%%%%%%%%%%%%%%%%%%%%%%%%%%%%%%

\myparagraph{Baselines}
\label{sec:baseline}
We choose the state-of-the-art method as a baseline in each dataset. We compared against DyNeRF, NeRFPlayer, MixVoxels, K-Planes, HyperReel, Nerfies, HyperNeRF, and TiNeuVox on the NeRF baseline. In detail, we use the version of NeRFPlayer TensoRF VM, HyperNeRF DF, Mixvoxels-L, K-Planes hybrid. We compared with 4DGaussians, 4DGS, and D3DGS based on the Gaussian baseline. Meanwhile, we have not included STG in our comparison due to its requirement for per-frame Structure from Motion (SfM) points, which makes conducting a fair comparison challenging. Also, STG is not a deformable 3D Gaussian approach. We followed the official code and configuration, except for increasing the training iterations for the Technicolor dataset to 1.5 times that of the 4DGaussians, in comparison to the Neural 3D Video dataset.

%%%%%%%%%%%%%%%%%%%%%%%%%%%%%%%%%%%%%%%%%%%%%%%%

\myparagraph{Metrics}
\label{sec:metrics}
We report the quality of rendered images using PSNR, SSIM, and LPIPS. Peak Signal-to-Noise Ratio (PSNR) quantifies pixel color error between the rendered video and the ground truth. We utilize SSIM \cite{wang2004image} to account for the perceived similarity of the rendered image. Additionally, we measure higher-level perceptual similarity using Learned Perceptual Image Patch Similarity (LPIPS) \cite{zhang2018unreasonable} with an AlexNet Backbone. Higher PSNR and SSIM values and lower LPIPS values indicate better visual quality.

%%%%%%%%%%%%%%%%%%%%%%%%%%%%%%%%%%%%%%%%%%%%%%%%

\begin{table}[tb]
\centering
\caption{\textbf{Average performance in the test view on  Neural 3D Video dataset} The computational cost was measured based on \texttt{flame\_salmon\_1} on the A6000. $^1$flame salmon scene includes only the first segment, comprising 300 frames. $^2$reported time and DyNeRF is trained on 8 GPUs and tested only on flame salmon. $^3$trained with 90 frames. $^4$trained with 50 frames.}
\vspace{-2mm}
\resizebox{0.8\textwidth}{!}{
\begin{tabular}{l|cccccc}
\multicolumn{1}{l|}{\multirow{2}{*}{model}} & \multicolumn{3}{c|}{metric}    & \multicolumn{3}{c}{computational cost} \\ \cline{2-7}
           & PSNR$\uparrow$ & SSIM$\uparrow$ & \multicolumn{1}{c|}{LPIPS$\downarrow$} & Training time$\downarrow$  & FPS$\uparrow$ & Model size$\downarrow$ \\ \hline
DyNeRF$^1$$^2$ \cite{li2022neural}                                                            & 29.58 & -     & \multicolumn{1}{c|}{0.083}  & 1344 hours      & 0.01 & 56 MB       \\
NeRFPlayer$^1$$^3$ \cite{song2023nerfplayer}                                                        & 30.69 & -     & \multicolumn{1}{c|}{0.111}  & 6 hours         & 0.05 & 1654 MB     \\
MixVoxels \cite{wang2022mixed}                                                        & 30.30 & 0.918 & \multicolumn{1}{c|}{0.127}  & 1 hours 40 mins & 0.93 & 512 MB      \\
K-Planes \cite{fridovich2023k}                                                          & 30.86 & 0.939 & \multicolumn{1}{c|}{0.096}  & 1 hours 40 mins & 0.13 & 309 MB      \\
HyperReel$^4$ \cite{attal2023hyperreel}                                                         & 30.37 & 0.921 & \multicolumn{1}{c|}{0.106}  & 9 hours 20 mins & 1.04  & 1362 MB     \\ \hline 
4DGS \cite{yang2023gs4d}                                                              & 31.19 & 0.940 & \multicolumn{1}{c|}{0.051}  & 9 hours 30 mins & 33.7 & 8700 MB     \\
4DGaussians \cite{wu20234dgaussians}                                                       & 30.71 & 0.935 & \multicolumn{1}{c|}{0.056}  & 50 mins         & 51.9 & 59 MB       \\
\textbf{Ours}                                                              & \textbf{31.31} & \textbf{0.945} & \multicolumn{1}{c|}{\textbf{0.037}}  & 1 hours 52 mins & 74.5 & 35 MB     
\end{tabular}
}
\label{tab:n3v}
\end{table}

\begin{figure}[tb]
\centering
\includegraphics[width=1\linewidth]{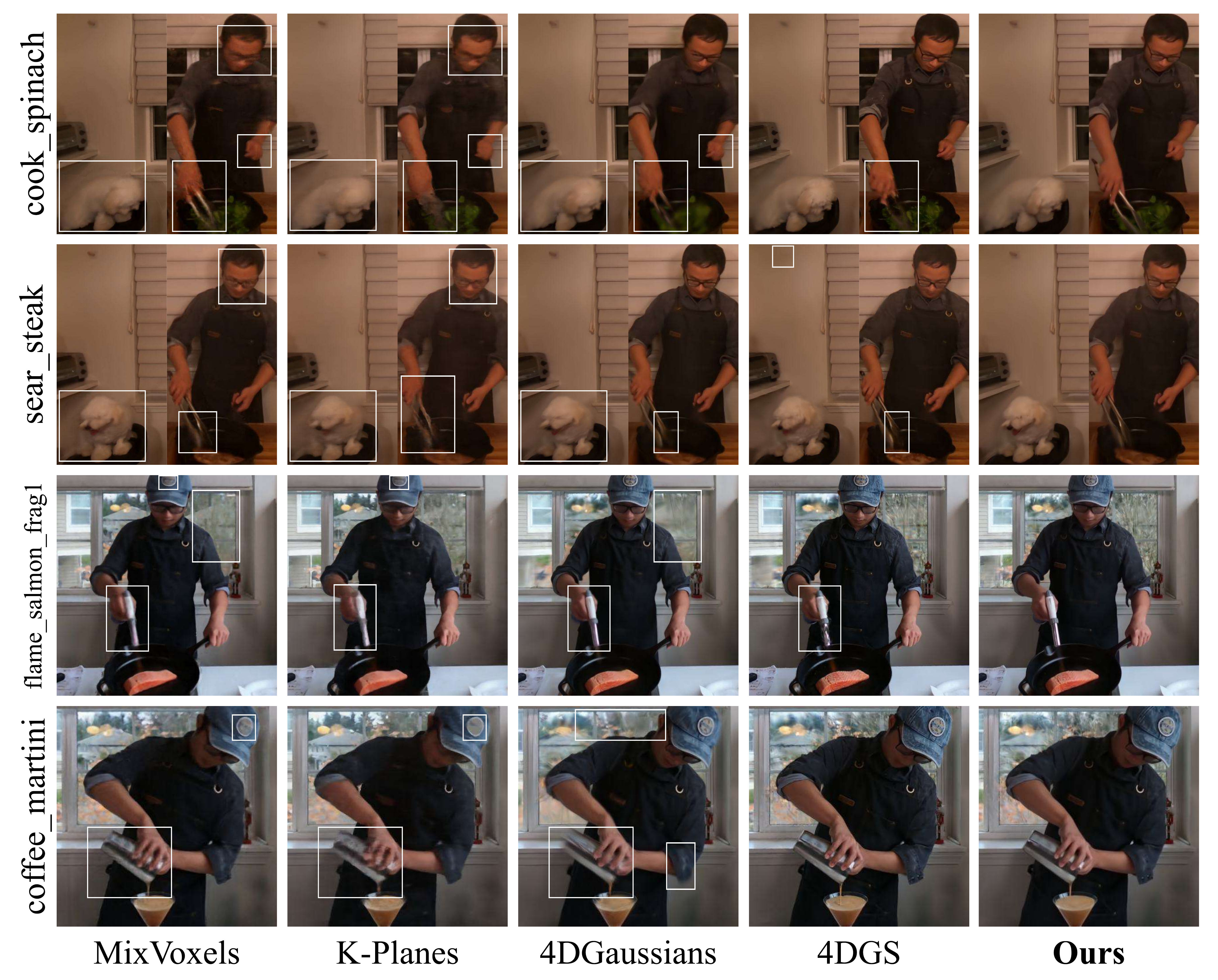}
\caption{\textbf{Qualitative comparisons on the  Neural 3D Video dataset.}} %The baseline models exhibit blurriness or artifacts in dynamic areas such as hands and torches, whereas ours successfully reconstructs them.}
\label{fig:n3v}
\end{figure}

\subsection{Effectiveness on Dynamic Region}
\label{sec:dynerf}

\myparagraph{Neural 3D Video Dataset}
 \cite{li2022neural} includes 20 multi-view videos, with each scene consisting of either 300 frames, except for the \texttt{flame\_salmon} scene, which comprises 1200 frames. These scenes encompass a relatively long duration and various movements, with some featuring multiple objects in motion. We utilized the Neural 3D Video dataset to observe the capability to capture dynamic areas. Total six scenes (\texttt{coffee\_martini, cook\_spinach, cut\_roasted\_beef, flame\_salmon, flame\_steak, sear\_steak}) are evaluated in \fref{fig:n3v} and \tref{tab:n3v}. The \texttt{flame\_salmon} scene is divided into four segments, each containing 300 frames. 

\tref{tab:n3v} presents quantitative metrics on the average metrics across the test views of all scenes, computational and storage costs on the first fragment of \texttt{flame\_salmon} scene. Refer to the Appendix for per-scene details. Our method demonstrates superior reconstruction quality, FPS, and model size across compared to baselines. As the table shows, NeRF baselines generally required longer training and rendering times. While 4DGS shows relatively high reconstruction performance, it demands longer training times and larger VRAM storage capacity compared to other baselines. 4DGaussians requires lower computational and storage costs but it displays low reconstruction quality in some scenes with rapid dynamics, as shown in the teaser and \fref{fig:n3v}. 

\fref{fig:n3v} reports the rendering quality. Our method successfully reconstructs the fine details in moving areas, outperforming baselines on average metrics across test views. Baselines show blurred dynamic areas or severe artifacts in low-light scenes such as \texttt{cook\_spinach} and \texttt{flame\_steak}. 
4DGS exhibits the disappearance of some static areas. In 4DGaussians, a consistent over-smoothing occurs in dynamic areas. All baselines experienced reduced quality in reflective or thin dynamic areas like clamps, torches, and windows.

%%%%%%%%%%%%%%%%%%%%%%%%%%%%%%%%%%%%%%%%%%%%%%%%

\myparagraph{Technicolor Light Field Dataset}
\label{sec:technicolor}

% \FloatBarrier
\begin{table}[ht]
\caption{\textbf{Average performance in the test view on Technicolor dataset}} 
\centering
\resizebox{0.7\textwidth}{!}{
\begin{tabular}{l|ccc|ccc}
%\multicolumn{7}{c}{HyperNeRF dataset} \\ \hline
\multicolumn{1}{l|}{\multirow{2}{*}{model}} & \multicolumn{3}{c|}{metric}    & \multicolumn{3}{c}{computational cost} \\ \cline{2-7}
 & {PSNR$\uparrow$} & {SSIM$\uparrow$} & {LPIPS$\downarrow$} & {Training time$\downarrow$} & {FPS$\uparrow$} & {Model size$\downarrow$} \\ \hline
DyNeRF                   & 31.80 & -   & 0.140 & -    & 0.02 & 0.6 MB  \\
HyperReel                 & 32.32 & 0.899   & 0.118 & 2 hours 45 mins & 0.91 & 289 MB  \\ \hline
4DGaussians              & 29.62 & 0.844   & 0.176 & 25 mins & 34.8 & 51 MB  \\
\textbf{Ours}                     & \textbf{33.24} & \textbf{0.907}   & \textbf{0.100} & 2 hours 55 mins & 60.8 & 77 MB  \\
\end{tabular}
}
\label{tab:technicolor_main}
\end{table}

\begin{figure}[ht]
\centering
\includegraphics[width=0.86\linewidth]{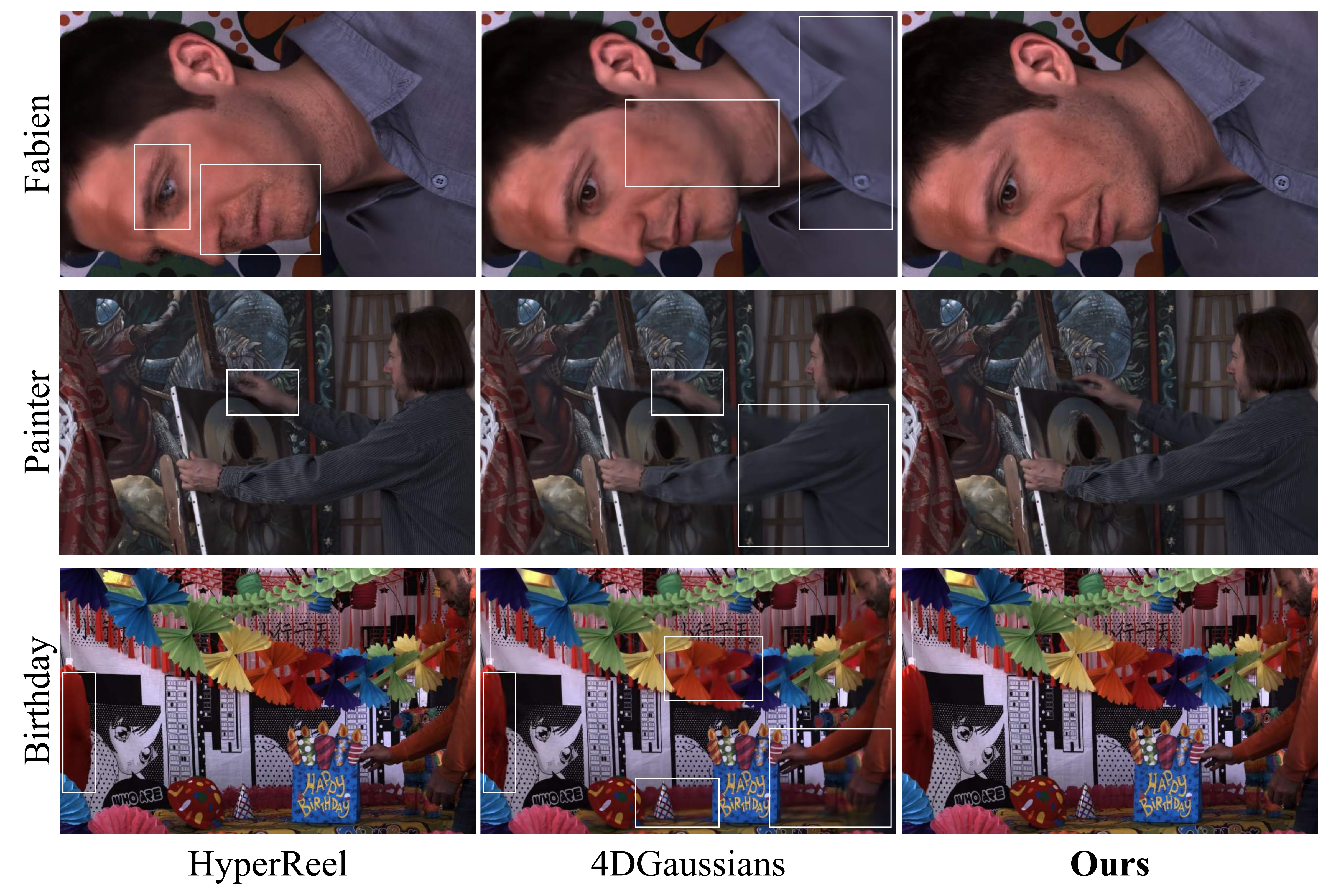}
\caption{\textbf{Qualitative comparisons on the Technicolor dataset.}}
\label{fig:technicolor}
\end{figure}

\cite{sabater2017dataset} is a multi-view dataset captured with a time-synchronized $4\times 4$ camera rig, containing intricate details. We train ours and the baselines on 50 frames of five commonly used scenes (\texttt{Birthday}, \texttt{Fabien}, \texttt{Painter}, \texttt{Theater}, \texttt{Trains}) using full-resolution videos at $2048 \times 1088$ pixels, with the second row and second column cameras used as test views.

\tref{tab:technicolor_main} reports the average metrics across the test views of all scenes, computational and storage costs on \texttt{Painter} scene. HyperReel demonstrates overall high-quality results but struggles with relatively slow training times and FPS, and a larger model size. 4DGaussians exhibits fast training times and FPS but significantly underperforms in reconstructing fine details compared to other baselines. However, our method demonstrates superior reconstruction quality and faster FPS compared to the baselines.

As shown in \fref{fig:technicolor}, HyperReel produces noisy artifacts due to incorrect predictions of the displacement vector. 4DGaussians fails to capture fine details in dynamic areas, exhibiting over-smoothing results. All baselines struggle to accurately reconstruct rapidly moving thin areas like fingers.

%%%%%%%%%%%%%%%%%%%%%%%%%%%%%%%%%%%%%%%%%%%%%%%%

\subsection{Challenging Camera Setting}
\label{sec:hypernerf}

% \FloatBarrier
\begin{table}[ht]
\caption{\textbf{Average performance in the test view on  Hypernerf dataset}}
\centering
\resizebox{0.7\textwidth}{!}{
\begin{tabular}{l|ccc|ccc}
\multicolumn{1}{l|}{\multirow{2}{*}{model}} & \multicolumn{3}{c|}{metric}    & \multicolumn{3}{c}{computational cost} \\ \cline{2-7}
 & {PSNR$\uparrow$} & {SSIM$\uparrow$} & {LPIPS$\downarrow$} & {Training time$\downarrow$} & {FPS$\uparrow$} & {Model size$\downarrow$} \\ \hline
Nerfies \cite{park2021nerfies}   & 22.23 & {-}     & 0.170 & $\sim$  hours & $<$ 1 & {-}    \\
HyperNeRF DS \cite{park2021hypernerf}             & 22.29 & 0.598   & \textbf{0.153} & 32 hours  & $<$ 1  & 15 MB  \\
TiNeuVox \cite{TiNeuVox}                 & 24.20 & 0.616   & 0.393 & 30 mins & 1 & 48 MB  \\ \hline
D3DGS \cite{yang2023deformable3dgs}                    & 22.40 & 0.598   & 0.275 & 3 hours 30 mins & 6.95 & 309 MB \\
4DGaussians              & 25.03 & 0.682   & 0.281 & 16 mins & 96.3 & 60 MB  \\
\textbf{Ours}                     & \textbf{25.43} & \textbf{0.697}   & 0.231 & 1 hours 15 mins & 139.3 & 33 MB  \\
\end{tabular}
}
\label{tab:hypernerf}

\end{table}

\begin{figure}[ht]
\centering
\includegraphics[width=0.7 \linewidth]{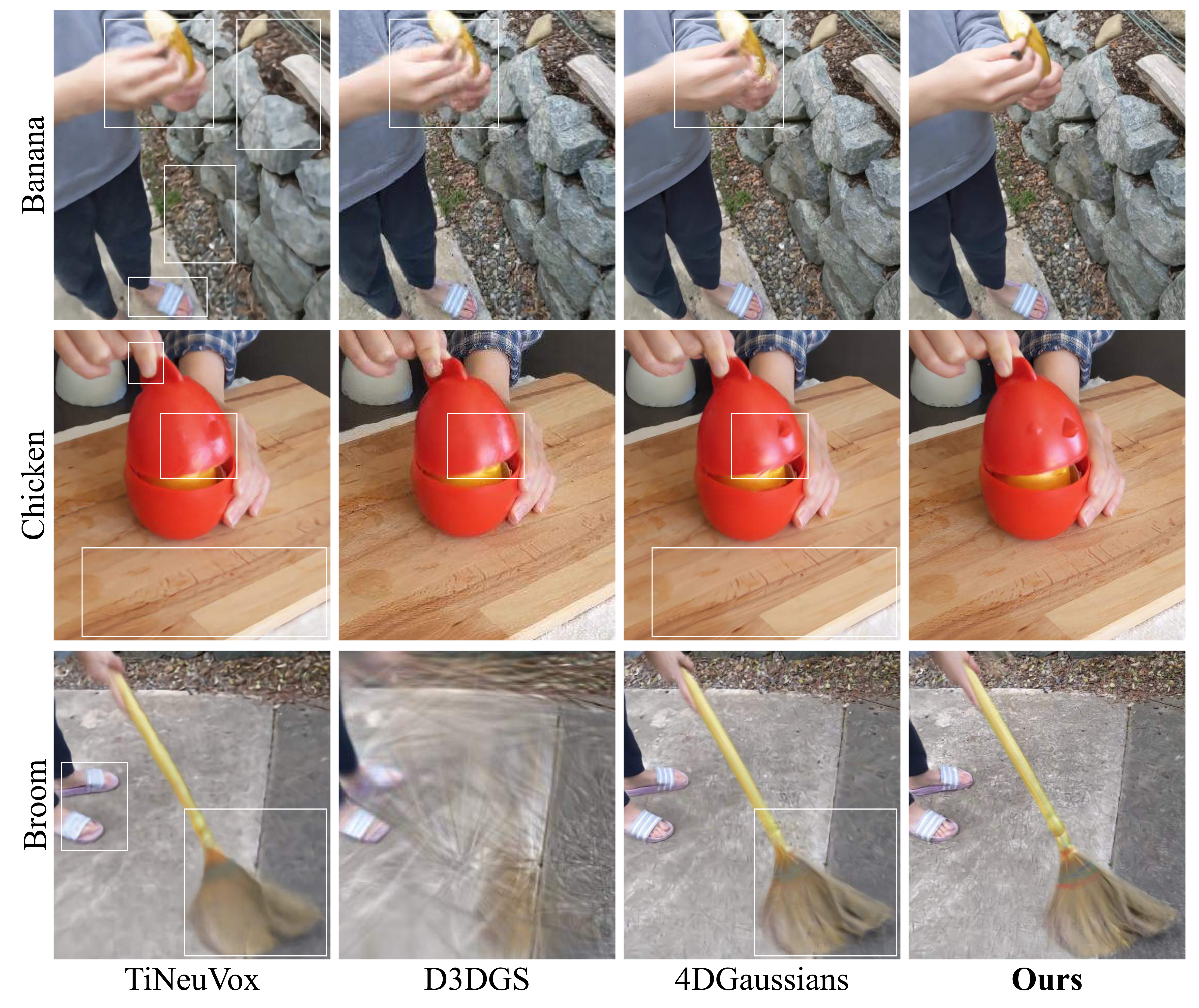}
\caption{\textbf{Qualitative comparisons on the HyperNeRF dataset.}}
\label{fig:hypernerf}
\end{figure}

\myparagraph{HyperNeRF Dataset}
includes videos captured using two phones rigidly mounted on a handheld rig. We train on all frames of four scenes (\texttt{3D Printer, Banana, Broom, Chicken}) at a resolution downsampled by half to $536 \times 960$. Due to memory constraints, D3DGS is trained on images downsampled by a quarter.

The table shows that our method outperforms the reconstruction performance with previous methods along with compact model size and faster FPS. \fref{fig:hypernerf} shows that previous methods struggle to reconstruct fast-moving parts such as fingers and broom. Especially D3DGS deteriorates in \texttt{Broom} scene. \tref{tab:hypernerf} reports the average metrics across the test views of all scenes, computational and storage costs on \texttt{Broom} scene. 

%%%%%%%%%%%%%%%%%%%%%%%%%%%%%%%%%%%%%%%%%%%%%%%%
%%%%%%%%%%%%%%%%%%%%%%%%%%%%%%%%%%%%%%%%%%%%%%%%
%%%%%%%%%%%%%%%%%%%%%%%%%%%%%%%%%%%%%%%%%%%%%%%%

\subsection{Analyses and Ablation Study}
\label{sec:analysis}

%%%%%%%%%%%%%%%%%%%%%%%%%%%%%%%%%%%%%%%%%%%%%%%%

\begin{figure}[ht]
\centering
\includegraphics[width=1\linewidth]{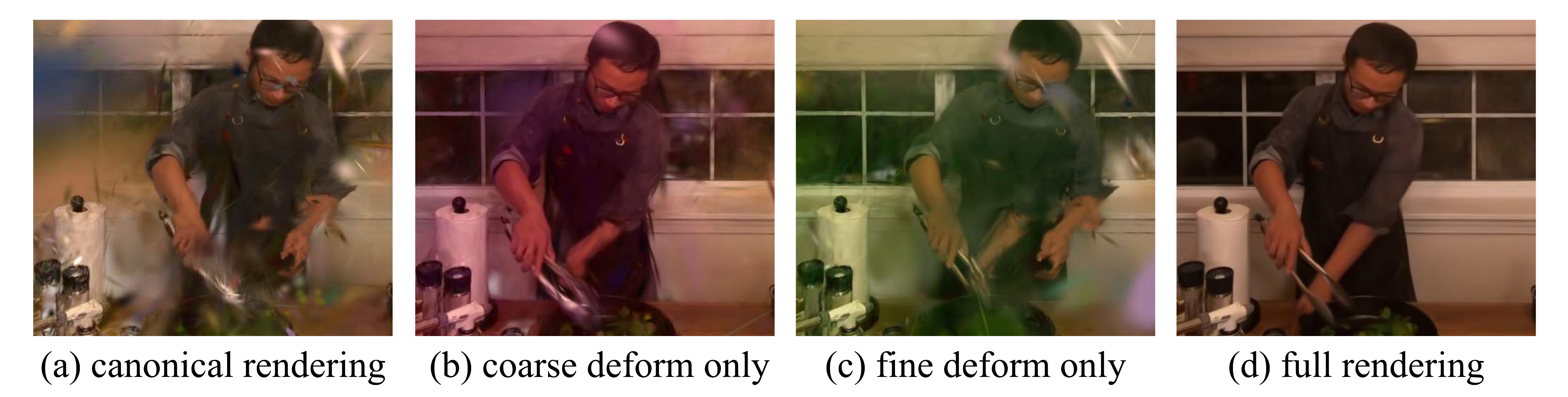}
\caption{\textbf{Deformation components.} (a) The canonical space contains Gaussians to represent all target frames of the scene. (b) Applying coarse deformation to the canonical space roughly reflects the dynamics of the scene. (c) The rendering without coarse deformation and only with fine deformation looks similar to the canonical rendering, i.e., responsible for fine deformations. (d) Applying both coarse and fine deformation yields natural rendering results.}
\label{fig:exp_deform}
\end{figure}

\begin{figure}[ht] 
\centering
\includegraphics[width=0.9 \linewidth]{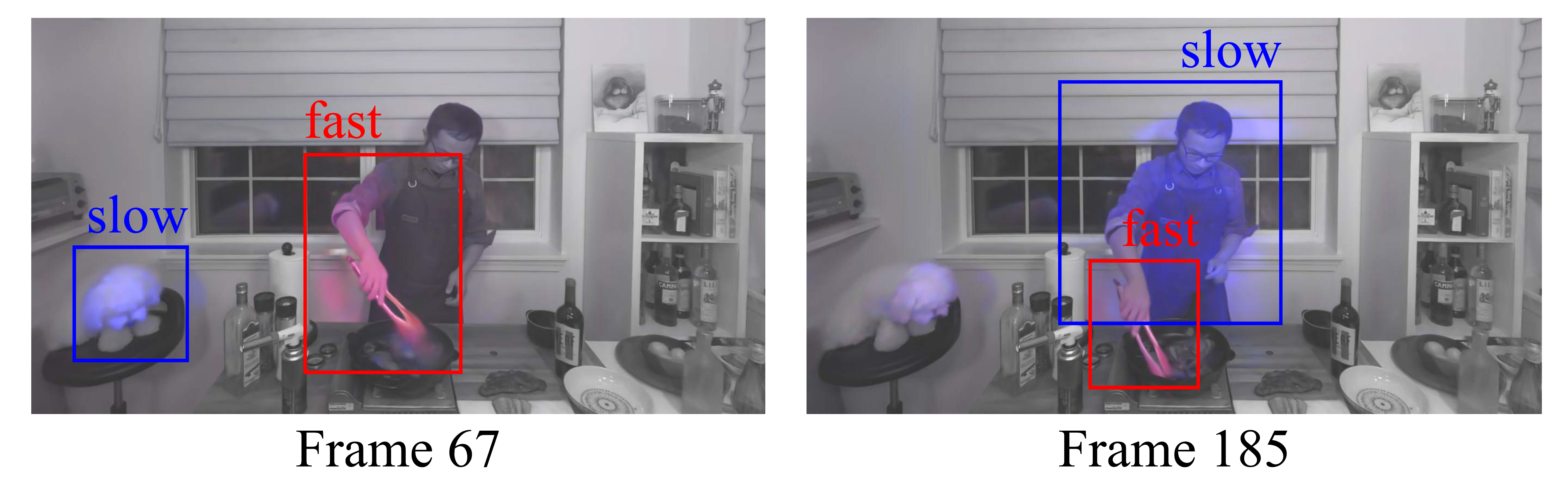}
\caption{\textbf{Visualization of the magnitude of deformation.} 
Coarse deformation (\textit{blue}) captures large and slow changes, such as the movement of the head and torso, while fine deformation (\textit{red}) is responsible for the fast and detailed movements of arms, tongs, shadows, \textit{etc}.}
\label{fig:supp_magnitude}
\end{figure}

\begin{figure}[tb] 
\centering
\includegraphics[width=1\linewidth]{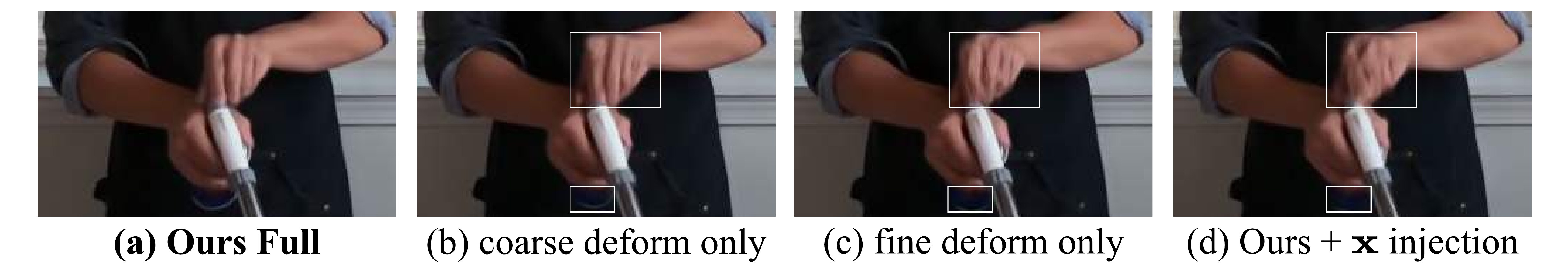}
\caption{\textbf{Qualitative ablation results on coarse-fine deformation.} (a) Our model achieves clear results with both coarse and fine decoders. (b-c) The quality of dynamic areas decreases if one is missing. (d) Additionally, introducing the coordinates of Gaussian as an additional input into our decoders results in a decrease in the quality of both static and dynamic regions.} 
\label{fig:supp_ablation}
\end{figure}

\myparagraph{Deformation components}
In \fref{fig:exp_deform}, we present an analysis of the coarse-fine deformation. To achieve this, we render a \texttt{flame\_steak} scene by omitting each of our deformation components one by one. Our full rendering results from adding coarse and fine deformation to the canonical space (\fref{fig:exp_deform}d). When both are removed, rendering yields an image in canonical space (\fref{fig:exp_deform}a). Rendering with the coarse deformation, which handles large or slow changes in the scene, produces results similar to the full rendering (\fref{fig:exp_deform}b). On the other hand, fine deformation is responsible for fast or detailed changes in the scene, yielding rendering similar to canonical space (\fref{fig:exp_deform}c).

To examine the roles of the coarse and fine deformation in the coarse-fine deformation, we conduct a visualization on \texttt{flame\_steak} scene. First, we compute the Euclidean norm of positional shifts between the current and subsequent frames. We then add the value to the DC components of the SH coefficients proportionally to the magnitude: blue for coarse deformation and red for fine deformation. For visual clarity, we render the original scene in grayscale. As illustrated in \fref{fig:supp_magnitude}, coarse deformation models slower changes such as body movement, while fine deformation models faster movements like cooking arms. Thus, we demonstrate that by downsampling the temporal embedding grid $Z$, we can effectively separate and model slow and fast deformations in the scene.

%%%%%%%%%%%%%%%%%%%%%%%%%%%%%%%%%%%%%%%%%%%%%%%%

% \FloatBarrier
\begin{table}[tb]
\caption{\textbf{Quantitative ablation results on coarse-fine deformation.}} % Training our model with only coarse/fine deformation or including center coordinates of Gaussian as additional inputs to the decoder results in degraded performance.}
\centering
\resizebox{0.7\textwidth}{!}{
\begin{tabular}{l|ccc}
Method & \;  PSNR$\uparrow$ \;  &  \; SSIM$\uparrow$ \;  &  \; LPIPS$\downarrow$ \;  \\ \hline
\textbf{Ours} & \textbf{29.70} & \textbf{0.933} & \textbf{0.041} \\ \hline
coarse deformation only & 29.48 & 0.931 & 0.044 \\
fine deformation only & 29.23 & 0.932  & 0.043 \\
Ours + \textbf{$\mathbf{x}$} injection & 29.60 & 0.931 & 0.045 \\
\end{tabular}
}
\label{tab:supp_ablation}
\end{table}

\myparagraph{Ablation study}

We report the results of an ablation study on the deformation decoder in \fref{fig:supp_ablation} and \tref{tab:supp_ablation}. First, our full method (using both coarse and fine decoders) produces clear rendering results and models dynamic variations well (\fref{fig:supp_ablation}a). Training only with the coarse or fine decoder leads to blurred dynamic areas and a failure to accurately capture detailed motion (\fref{fig:supp_ablation}b-c). Additionally, we demonstrate experiments where the Gaussian center coordinates $\pos$ are injected into the input of each decoder. As shown in \fref{fig:supp_ablation}d, including the Gaussian coordinates degrades the quality of deformation, supporting our argument that coordinate dependency should be removed from the deformation function.

\begin{table}[ht]
\centering
\caption{\textbf{Quantitative ablation results on local smoothness regularization.} We compare the performance of applying our regularization with the physically-based regularization of Dynamic 3D Gaussians \cite{luiten2023dynamic}. Our regularization better captures details of dynamic objects.}
\resizebox{0.75\textwidth}{!}{
\begin{tabular}{l|ccc}
Method & \;  PSNR$\uparrow$ \;  &  \; SSIM$\uparrow$ \;  &  \; LPIPS$\downarrow$ \;  \\ \hline
\textbf{Ours w/o embedding reg} & 32.26 & 0.951 & 0.037 \\ \hline
+ our embedding reg & \textbf{32.34} & \textbf{0.952} & \textbf{0.036} \\
+ physically-based reg & 32.08 & 0.950  & 0.036 \\
\end{tabular}
}
\label{tab:reg_ablation}
\end{table}

\begin{figure}[tb] 
\centering
\includegraphics[width=0.8\linewidth]{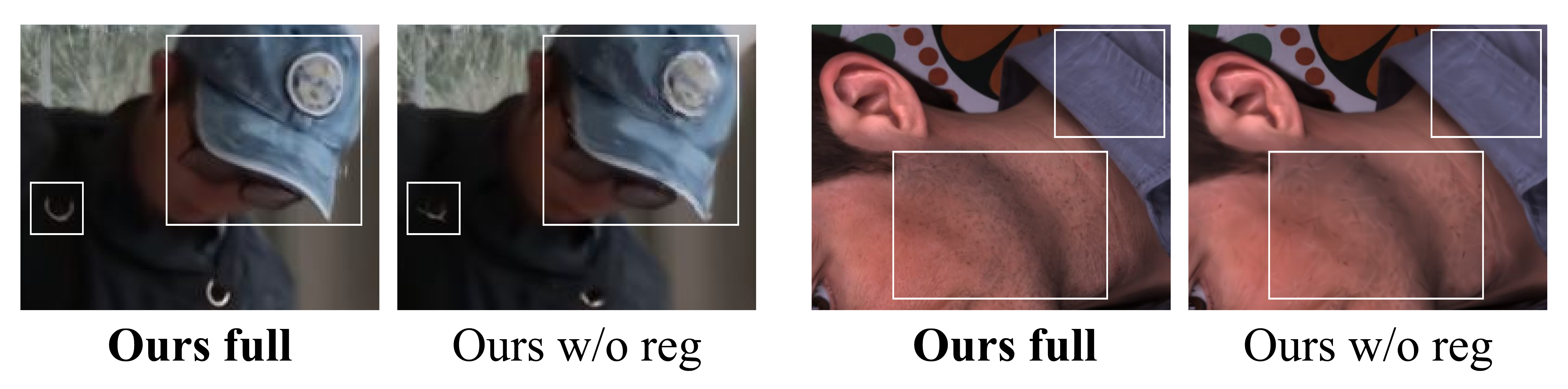}
\caption{\textbf{Qualitative ablation results on local smoothness regularization.}} 
\label{fig:reg_ablation}
\end{figure}

Furthermore, we report the results of an ablation study on the local smoothness regularization for per-Gaussian embeddings. As shown in \fref{fig:reg_ablation}, our regularization improves the details and texture quality of moving objects. In \tref{tab:reg_ablation}, we show a performance comparison between the proposed regularization and the existing physically-based regularizations. To apply the method proposed by Luiten et al.\cite{luiten2023dynamic} to ours, we make some modifications: 1) Like our method, we find the set of k-nearest neighbors only when the densification occurs to reduce the computational cost. 2) For long-term local-isometry loss, we use the time of the video frame used in the previous training step instead of using the time of the first frame. Our regularization is simple and shows better performance compared to the previous method.

\begin{figure}[tb]
\centering
\includegraphics[width=0.9\linewidth]{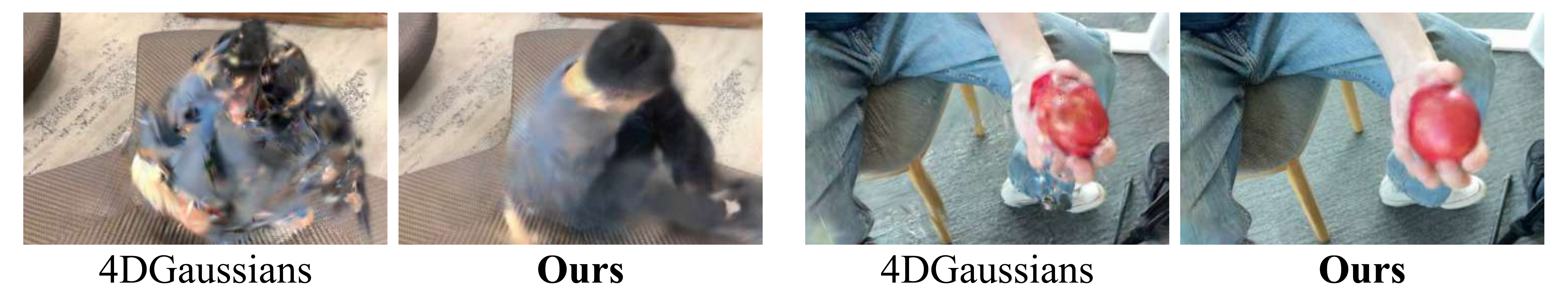}
\caption{\textbf{Limitation.} Ours struggles with the casually captured monocular videos.}
\label{fig:limitation}
\end{figure}

%%%%%%%%%%%%%%%%%%%%%%%%%%%%%%%%%%%%%%%%%%%%%%%%
%%%%%%%%%%%%%%%%%%%%%%%%%%%%%%%%%%%%%%%%%%%%%%%%
%%%%%%%%%%%%%%%%%%%%%%%%%%%%%%%%%%%%%%%%%%%%%%%%

\section{Conclusion and Limitation}
We propose a per-Gaussian deformation for 3DGS that takes per-Gaussian embeddings as input, instead of using the typical deformation fields from previous deformable 3DGS works, resulting in high performance. We enhance the reconstruction quality by decomposing the dynamic changes into coarse and fine deformation. However, our method learns inappropriate Gaussian deformation with casually captured monocular videos\cite{gao2022dynamic}, like other baselines. 
We plan to address it in future work by introducing useful prior for monocular video settings.

%%%%%%%%%%%%%%%%%%%%%%%%%%%%%%%%%%%%%%%%%%%%%%%%
%%%%%%%%%%%%%%%%%%%%%%%%%%%%%%%%%%%%%%%%%%%%%%%%
%%%%%%%%%%%%%%%%%%%%%%%%%%%%%%%%%%%%%%%%%%%%%%%%

\section*{Acknowledgements}
This work is supported by the Institute for Information \& Communications Technology Planning \& Evaluation (IITP) grant funded by the Korea government(MSIT) (No. 2017-0-00072, Development of Audio/Video Coding and Light Field Media Fundamental Technologies for Ultra Realistic Tera-media)

% \input{tabs/project_page}

\UseRawInputEncoding
\def\supp{1}  % comment this for the main paper submission
\ifx \supp \undefined
    %%%%%%%%% REFERENCES
    {
    \bibliographystyle{splncs04}
    \bibliography{main}
    }
\else
    \bibliographystyle{splncs04}
    \bibliography{main}
    \clearpage
    \ifx \supp \undefined
    \documentclass[runningheads]{llncs}
    
    % ---------------------------------------------------------------
    % Include basic ECCV package

    \usepackage{eccv}

    \usepackage{lipsum}
    \usepackage{ulem}
    \usepackage{multirow}
    \usepackage{hhline}
    \usepackage{placeins}
    \usepackage{float}
    \usepackage{colortbl} 
    % ---------------------------------------------------------------
    % Other packages
    
    % Commonly used abbreviations (\eg, \ie, \etc, \cf, \etal, etc.)
    \usepackage{eccvabbrv}
    \usepackage{caption} 
    % Include other packages here, before hyperref.
    \usepackage{graphicx}
    \usepackage{booktabs}
    
    % The "axessiblity" package can be found at: https://ctan.org/pkg/axessibility?lang=en
    \usepackage[accsupp]{axessibility}  % Improves PDF readability for those with disabilities.
    \usepackage{setspace}
    
    % ---------------------------------------------------------------
    % Hyperref package
    \usepackage[pagebackref,breaklinks,colorlinks,citecolor=eccvblue]{hyperref}
    % Support for ORCID icon
    \usepackage{orcidlink}
    \usepackage{tikzducks}
    \usepackage{graphicx}
    \usepackage[strict]{changepage}
    \usepackage{diagbox} 
    \usepackage{tabularx} % for adjustable-width columns
    \usepackage{adjustbox} % for resizing tables

    \begin{document}
\fi

\begin{appendix}

\renewcommand{\thetable}{S\arabic{table}}
\renewcommand{\thefigure}{S\arabic{figure}}
\setcounter{figure}{0}
\setcounter{table}{0}

In the Appendix, we provide additional training details (\aref{sup:details}) and more results of the baseline, along with the per-scene quantitative and qualitative results of our method on all datasets (\aref{sup:results}).

\section{Training Details}
\label{sup:details}

\subsection{Efficient Training Strategy}
\label{sec:training}

\begin{figure}[ht]
\centering
\includegraphics[width=1\linewidth]{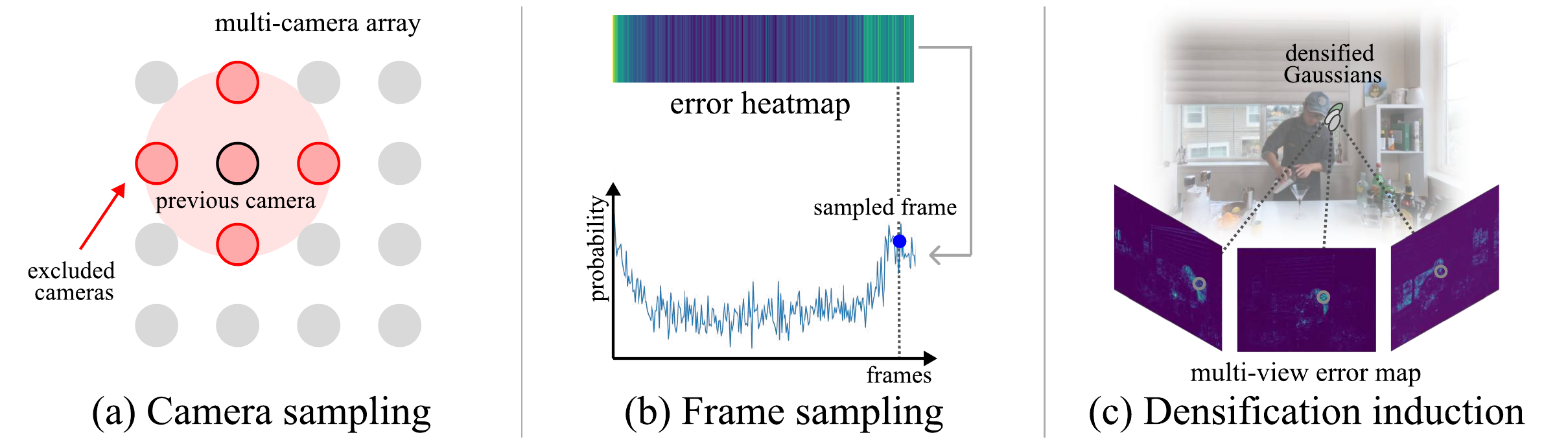}
\caption{\textbf{Components of efficient training strategy.}} 
\label{fig:supp_sampling}
\end{figure}

We introduce a training strategy for faster convergence and higher performance. The first is to evenly cover multi-view camera perspectives and exclude the camera index that was used in the previous iteration. We pre-compute pairwise distances of all camera origins before the start of the training. Then we exclude the cameras with distance to the previous camera less than 40 percentile of all distances.

The second is to sample frames within the target cameras. Different frames may have different difficulties for reconstruction. We measure the difficulty of a frame by the magnitude of error at the frame.  Then, we sample training frames with a categorical distribution which is proportional to the magnitude of error. For multi-view videos, we average the errors from all viewpoints. For the first 10K iterations, we randomly sample the frames, and then alternately use random sampling and error-based frame sampling.

The third is to change the policy for Gaussian densification. Previous methods invoke the densification by minimizing L1 loss \cite{wu20234dgaussians} or L1 loss and DSSIM loss for every iteration \cite{yang2023real,yang2023deformable3dgs,li2023spacetimegaussians}. We observe that the DSSIM loss improves the visual quality in the background but takes longer training time. Therefore, we use L1 loss for every iteration and periodically use the additional multi-view DSSIM loss in the frame with high error. This encourages densification in the region where the model struggles. We fix the frame obtained by loss-based sampling for every 50 iterations, and minimize the multi-view DSSIM loss through camera sampling during the next 5 iterations. Similar to frame sampling, DSSIM loss is applied after the first 10K iterations. 

\newpage
\subsection{Implementation Details}
\label{sup:imple_details}

% \FloatBarrier
\begin{table}[ht]
\centering
\caption{\textbf{Ablation study on the size of temporal and per-Gaussian embeddings.} We compare the results on the \texttt{flame\_steak} scene. We set 256 /32 as the default setting.}
\resizebox{\textwidth}{!}{
\begin{tabular}{c|ccc||c|ccc}
$\embt$ dim/$\embg$ dim & PSNR $\uparrow$ & SSIM $\uparrow$ & LPIPS $\downarrow$ & $\embt$ dim/$\embg$ dim & PSNR $\uparrow$ & SSIM $\uparrow$ & LPIPS $\downarrow$ \\
\hhline{-|---||-|---}
256 / 16 & 33.53  & 0.962 & 0.030  & 128 / 32 & \textbf{33.61} & 0.964 & 0.029 \\
256 / 32 & 33.57 & \textbf{0.964} & \textbf{0.028} & 256 / 32 & 33.57  & \textbf{0.964} & \textbf{0.028} \\
256 / 64 & \textbf{33.71} & 0.964 & 0.029 & 512 / 32 & 33.55 & 0.964 & 0.028 \\
%\hhline{-|---||-|---}
\end{tabular}
}
\label{tab:abl_dim}
\end{table}

We report the performance variation by changing the size of per-Gaussian and temporal embeddings in \tref{tab:abl_dim}. The embedding size does not significantly affect the overall performance. In all experiments, considering the capacity and details, we use a 32-dimensional vector for Gaussian embeddings $\embg$ and a 256-dimensional vector for temporal embeddings $\embt$. 

The decoder MLP consists 128 hidden units and the MLP head for Gaussian parameters are both composed of 2 layers with 128 hidden units.  For Technicolor and HyperNeRF datasets, we use a 1-layer decoder MLP. To efficiently and stabilize the initial training, we start the temporal embedding grid  $Z_t^{\text{f}}$ at the same $N/5$ time resolution as $Z_t^{\text{c}}$ and gradually increase it to $N$ resolution over 10K iterations, with $N$ set to 150 for 300 frames.
The learning rate of our deformation decoder starts at $1.6 \times 10^{-4}$ and exponentially decays to $1.6 \times 10^{-5}$ over the total training iterations. The learning rate of the temporal embedding $\embt$ follows that of the deformation decoder, while the learning rate of per-Gaussian embedding $\embg$ is set to $2.5 \times 10^{-3}$.
We eliminate the opacity reset step of 3DGS and instead add a loss that minimizes the mean of Gaussian opacities with a weight of $1.0\times 10^{-4}$. We empirically set the start of the training strategy to 10K iterations. 

For the Neural 3D Video dataset, we follow Sync-NeRF\cite{kim2023sync} to introduce trainable time offsets to each camera. In addition, we learn the deformation of color, scale, and opacity from 5K iterations and perform densification and pruning.

For initialization points, we downsample point clouds obtained from COLMAP dense scene reconstruction. For the Neural 3D Video dataset, we train for 80K iterations; for the Technicolor dataset, 100K for the \texttt{Birthday} and \texttt{Painter} scenes and 120K for the rest; for the HyperNeRF dataset, 80K for the \texttt{Banana} scene and 60K for the others. We periodically use the DSSIM loss at specific steps, excluding the HyperNeRF dataset. %For details, please refer to the Appendix.
For calculating computational cost, the Neural 3D Video dataset was measured using A6000, while the Technicolor Light Field dataset and the HyperNeRF dataset were measured using 3090.

For local smoothness embedding regularization, we set the weight of loss to 1 on the Neural 3D Video dataset and 0.1 on the Technicolor Light Field dataset and the HyperNeRF dataset.

%%%%%%%%%%%%%%%%%%%%%%%%%%%%%%%%%%%%%%%%%%%%%%%%
%%%%%%%%%%%%%%%%%%%%%%%%%%%%%%%%%%%%%%%%%%%%%%%%
%%%%%%%%%%%%%%%%%%%%%%%%%%%%%%%%%%%%%%%%%%%%%%%%
\newpage
\section{More Results}
\label{sup:results}
\begin{figure}[ht]
\centering
\includegraphics[width=1\linewidth]{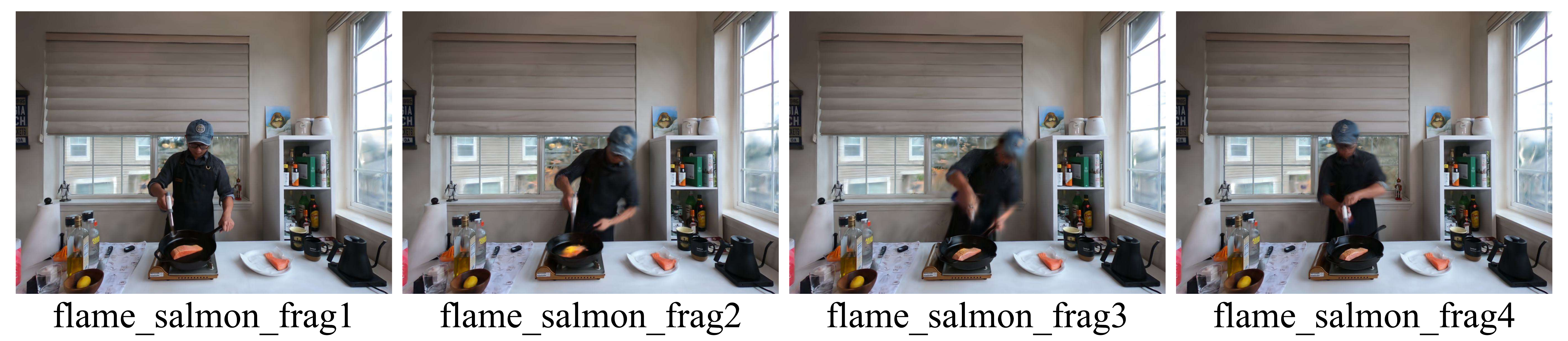}
\caption{\textbf{Rendering results of 4DGaussians on Neural 3D Video dataset.}}
\label{fig:supp_4dgs}
\end{figure}

\begin{figure}[H] 
% \begin{figure}[tb]
\centering
\includegraphics[width=1\linewidth]{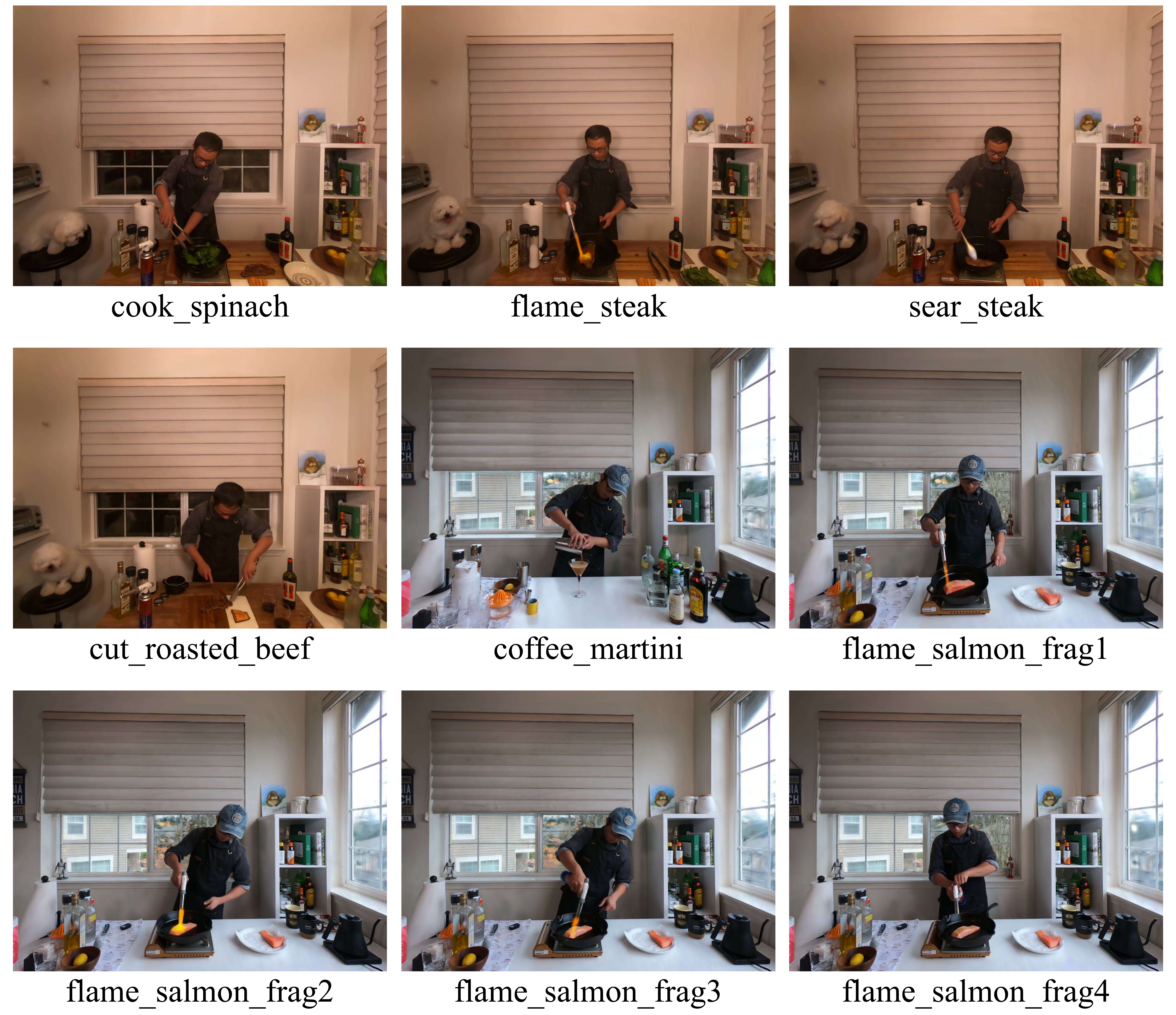}
\caption{\textbf{Rendering results on Neural 3D Video dataset.}}
\label{fig:supp_dynerf}
\end{figure}

\begin{figure}[H] 
\centering
\includegraphics[width=1\linewidth]{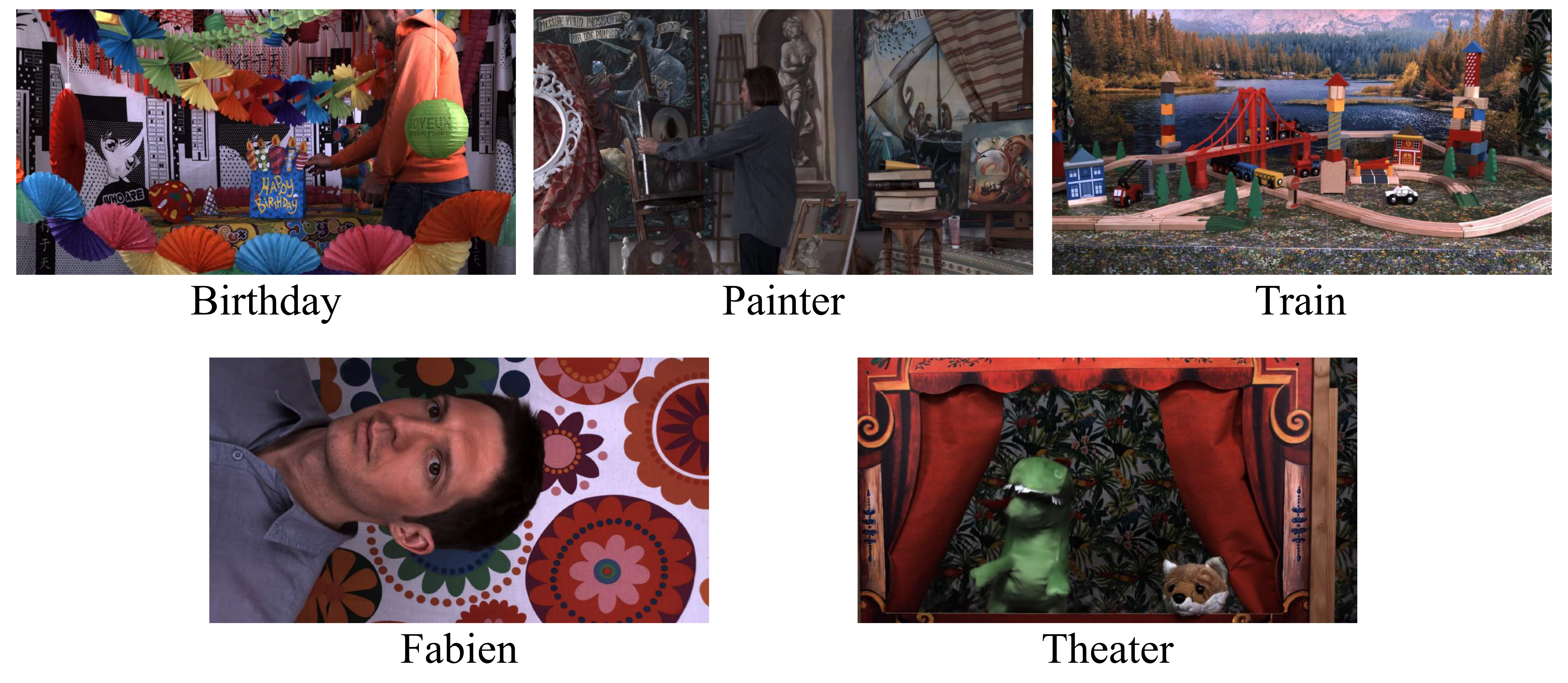}
\caption{\textbf{Rendering results on Technicolor Light Field dataset.}}
\label{fig:supp_technicolor}
\end{figure}

\begin{figure}[H] 
\centering
\includegraphics[width=1\linewidth]{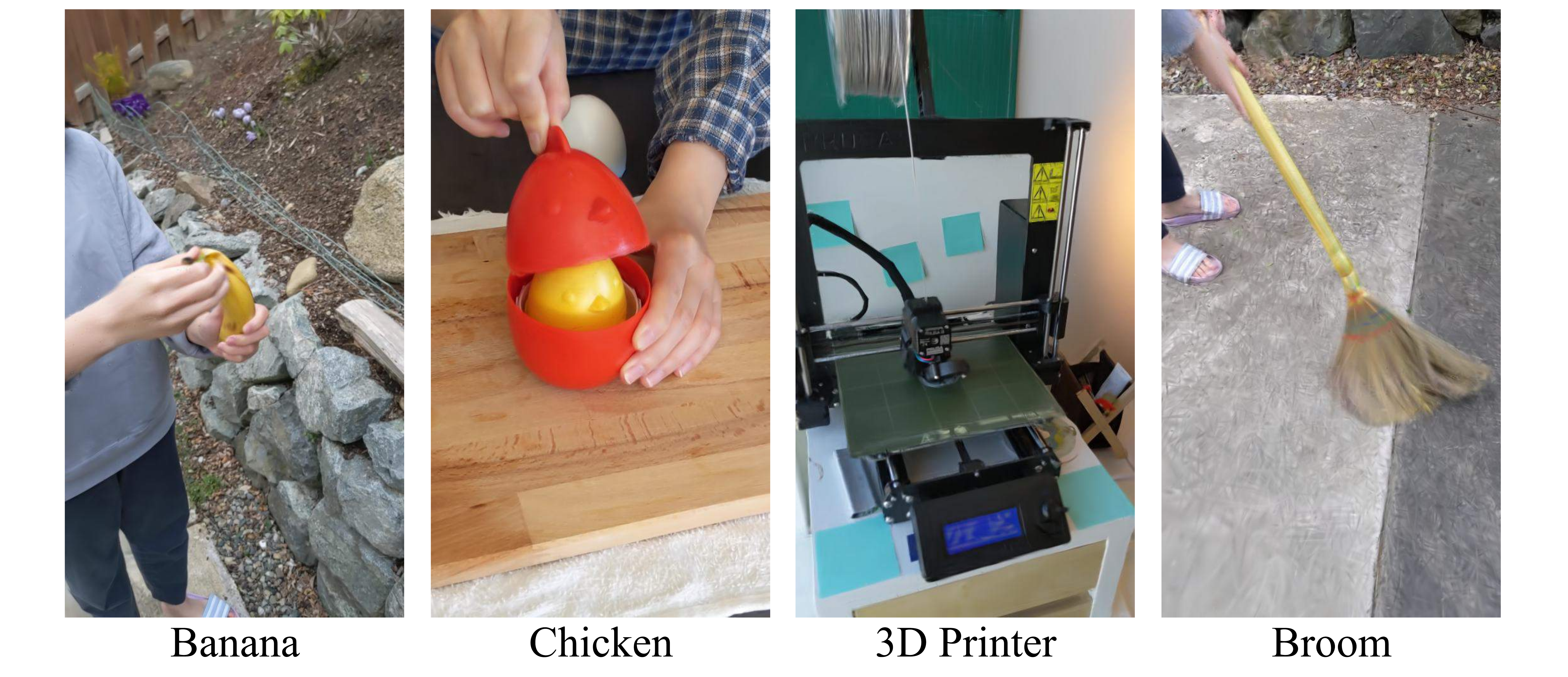}
\caption{\textbf{Rendering results on HyperNeRF dataset.}}
\label{fig:supp_hypernerf}
\end{figure}

We show the additional rendering of 4DGasussians \cite{wu20234dgaussians} in \fref{fig:supp_4dgs}. And, we show the entire rendering results of ours in \fref{fig:supp_dynerf}-\ref{fig:supp_hypernerf}. In most scenes, our model shows better perceptual performance than the baselines. \tref{tab:supp_n3v}-\ref{tab:supp_hypernerf} report the quantitative results of each scene on the Neural 3D Video dataset, Technicolor Light Field dataset, and HyperNeRF dataset.

\subsection{Additional Results of 4DGaussians}

The results of 4DGaussians are produced using the official repository. In the teaser, we choose the \texttt{third} fragment of \texttt{flame salmon} scene which contains more difficult movements than the typical first fragment to demonstrate our superiority.

\newpage
\subsection{Per-Scene Results of Neural 3D Video Dataset}
\FloatBarrier

% \FloatBarrier
\begin{table}[H]
\caption{\textbf{Per-scene quantitative results on Neural 3D Video dataset}}
\centering
\resizebox{0.85\textwidth}{!}{%
\begin{tabular}{l|cc|cc|cc|cc|cc}
\multicolumn{1}{l|}{\multirow{2}{*}{\diagbox[]{Model}{Metric}}} & \multicolumn{2}{c|}{Average} & \multicolumn{2}{c|}{\texttt{coffee\_martini}} & \multicolumn{2}{c|}{\texttt{cook\_spinach}} & \multicolumn{2}{c|}{\texttt{cut\_roasted\_beef}} & \multicolumn{2}{c}{\texttt{flame\_salmon\_1}} \\ \cline{2-11}
               & PSNR$\uparrow$      & SSIM$\uparrow$      & PSNR$\uparrow$         & SSIM$\uparrow$         & PSNR$\uparrow$         & SSIM$\uparrow$        & PSNR$\uparrow$        & SSIM$\uparrow$       & PSNR$\uparrow$            & SSIM$\uparrow$           \\ \hline
MixVoxels      & 30.30       & 0.918        & \second 29.44          & 0.916           & 29.97          & 0.934          & 32.58         & 0.938         & \second 30.50             & 0.918             \\
K-Planes       & 30.86       & 0.939        & \best 29.66          & \second 0.926           & 31.82          & 0.943          & 31.82         & \best 0.966         & \best 30.68             & 0.928             \\
HyperReel  & 30.37       & 0.921        & 28.37          & 0.892           & 32.30          & 0.941          & 32.92         & 0.945         & 28.26             & 0.882             \\ \hline
4DGS    & \second 31.19       & \second 0.940        & 28.63          & 0.918           & \best 33.54          & \second 0.956          & \best 34.18         & \second 0.959         & 29.25             & \second 0.929             \\ 
4DGaussians           & 30.71       & 0.935        & 28.44          & 0.919           & \second 33.10          & 0.953          & 33.32         & 0.954         & 28.80             & 0.926             \\
\textbf{Ours}           & \best 31.31       & \best 0.945        & 29.10          & \best 0.931           & 32.96          & \best 0.956          & \second 33.57         & 0.958         & 29.61             & \best 0.936             \\ \hline \hline

\multicolumn{1}{l|}{\multirow{2}{*}{\diagbox[]{Model}{Metric}}} & \multicolumn{2}{c|}{\texttt{flame\_salmon\_2}} & \multicolumn{2}{c|}{\texttt{flame\_salmon\_3}} & \multicolumn{2}{c|}{\texttt{flame\_salmon\_4}} & \multicolumn{2}{c|}{\texttt{flame\_steak}} & \multicolumn{2}{c}{\texttt{sear\_steak}} \\ \cline{2-11}
               & PSNR$\uparrow$      & SSIM$\uparrow$      & PSNR$\uparrow$         & SSIM$\uparrow$         & PSNR$\uparrow$         & SSIM$\uparrow$        & PSNR$\uparrow$        & SSIM$\uparrow$       & PSNR$\uparrow$        & SSIM$\uparrow$       \\ \hline
MixVoxels      & \best 30.53       & 0.915        & 27.83          & 0.853           & 29.49          & 0.899          & 30.74         & 0.945         & 31.61         & 0.949         \\
K-Planes       & \second 29.98       & 0.924        & \best 30.10          & 0.924           & \best 30.37          & 0.922          & 31.85         & \best 0.969         & 31.48         & 0.951         \\
HyperReel  & 28.80       & 0.911        & 28.97          & 0.911           & 28.92          & 0.908          & 32.20         & 0.949         & 32.57         & 0.952         \\ \hline
4DGS    & 29.28       & \second 0.929        & 29.40          & \second 0.927           & 29.13          & \second 0.925          & \best 33.90         & 0.961         & 33.37         & 0.960         \\ 
4DGaussians           & 28.21       & 0.913        & 28.68          & 0.917           & 28.45          & 0.913          & 33.55         & 0.961         & \best 34.02         & \second 0.963         \\
\textbf{Ours}           & 29.69       & \best 0.934        & \second 29.94          & \best 0.934           & \second 29.90          & \best 0.933          & \second 33.57         & \second 0.964         & \second 33.45         & \best 0.963         \\ 
\end{tabular}%
}

\label{tab:supp_n3v}
\end{table}

%%%%%%%%%%%%%%%%%%%%%%%%%%%%%%%%%%%%%%%%%%%%%%%%

\newpage
\subsection{Per-Scene Results of Technicolor Light Field Dataset}

\FloatBarrier

% \FloatBarrier
\begin{table}[H]
\centering
\caption{\textbf{Per-scene quantitative results on Technicolor dataset}}
\resizebox{0.65\textwidth}{!}{
\begin{tabular}{lcccccc}
\multicolumn{1}{l|}{\multirow{2}{*}{\diagbox[]{Model}{Metric}}} & \multicolumn{2}{c|}{Average}       & \multicolumn{2}{c|}{\texttt{Birthday}}      & \multicolumn{2}{c}{\texttt{Fabien}} \\ \cline{2-7} 
\multicolumn{1}{l|}{}                                             & PSNR$\uparrow$  & \multicolumn{1}{c|}{SSIM$\uparrow$}  & PSNR$\uparrow$  & \multicolumn{1}{c|}{SSIM$\uparrow$}  & PSNR$\uparrow$         & SSIM$\uparrow$        \\ \hline

\multicolumn{1}{l|}{DyNeRF}                                       & 31.80 & \multicolumn{1}{c|}{-}     & 29.20 & \multicolumn{1}{c|}{-}     & 32.76        & -           \\

\multicolumn{1}{l|}{HyperReel}                                    & \second 32.32 & \multicolumn{1}{c|}{\second 0.899} & \second 30.57 & \multicolumn{1}{c|}{\second 0.918} & 32.49        & 0.863       \\ \hline

\multicolumn{1}{l|}{4DGaussians}                                  & 29.62 & \multicolumn{1}{c|}{0.840} & 28.03 & \multicolumn{1}{c|}{0.862} & \second 33.36        & \second 0.865       \\ 

\multicolumn{1}{l|}{\textbf{Ours}} & \best 33.23 & \multicolumn{1}{c|}{\best 0.907} & \best 32.90 & \multicolumn{1}{c|}{\best 0.951} & \best 34.71 & \best 0.885 \\ \hline \hline

\multicolumn{1}{l|}{\multirow{2}{*}{\diagbox[]{Model}{Metric}}} & \multicolumn{2}{c|}{\texttt{Painter}} & \multicolumn{2}{c|}{\texttt{Theater}}       & \multicolumn{2}{c}{\texttt{Train}}  \\ \cline{2-7} 

\multicolumn{1}{l|}{}                                             & PSNR$\uparrow$  & \multicolumn{1}{c|}{SSIM$\uparrow$}  & PSNR$\uparrow$  & \multicolumn{1}{c|}{SSIM$\uparrow$}  & PSNR$\uparrow$         & SSIM$\uparrow$        \\ \hline

\multicolumn{1}{l|}{DyNeRF} & \second 35.95 & \multicolumn{1}{c|}{-}     & 29.53 & \multicolumn{1}{c|}{-} & \best 31.58 & -  \\

\multicolumn{1}{l|}{HyperReel} & 35.51 & \multicolumn{1}{c|}{\best 0.924} & \best 33.76 & \multicolumn{1}{c|}{\best 0.897} & 29.30 & \second 0.894 \\ \hline

\multicolumn{1}{l|}{4DGaussians} & 34.52 & \multicolumn{1}{c|}{0.899} & 28.67 & \multicolumn{1}{c|}{0.835} & 23.54 & 0.756 \\ 

\multicolumn{1}{l|}{\textbf{Ours}} & \best 36.18 & \multicolumn{1}{c|}{\second 0.924} & \second 31.07 & \multicolumn{1}{c|}{\second 0.868} & \second 31.33 & \best 0.912      
\end{tabular}
}
\label{tab:supp_technicolor}
\end{table}

%%%%%%%%%%%%%%%%%%%%%%%%%%%%%%%%%%%%%%%%%%%%%%%%

\newpage
\subsection{Per-Scene Results of HyperNeRF Dataset}

\FloatBarrier

% \FloatBarrier
\begin{table}[H] 
\centering
\caption{\textbf{Per-scene quantitative results on HyperNeRF dataset}}
\resizebox{\textwidth}{!}{
\begin{tabular}{l|cc|cc|cc|cc|cc}
\multicolumn{1}{l|}{\multirow{2}{*}{\diagbox[]{Model}{Metric}}} & \multicolumn{2}{c|}{Average}        & \multicolumn{2}{c|}{\texttt{Broom}} & \multicolumn{2}{c|}{\texttt{3D Printer}}    & \multicolumn{2}{c|}{\texttt{Chicken}}                                                & \multicolumn{2}{c}{\texttt{Banana}} \\ \cline{2-11}  
\multicolumn{1}{l|}{} & PSNR$\uparrow$  & \multicolumn{1}{c|}{SSIM$\uparrow$}   & PSNR$\uparrow$         & \multicolumn{1}{c|}{SSIM$\uparrow$} & PSNR$\uparrow$  & \multicolumn{1}{c|}{SSIM$\uparrow$}  & PSNR$\uparrow$                            & \multicolumn{1}{c|}{SSIM$\uparrow$}                 & PSNR$\uparrow$         & SSIM$\uparrow$               \\ \hline

\multicolumn{1}{l|}{Nerfies}                                      & 22.23 & \multicolumn{1}{c|}{-}     &  19.30         & -      & 20.00 & \multicolumn{1}{c|}{-}     & 26.90                           & \multicolumn{1}{c|}{-}                    & 23.30        & -                \\

\multicolumn{1}{l|}{HyperNeRF DS}                                 & 22.29 & \multicolumn{1}{c|}{0.598} & 19.51        & 0.210    & 20.04 & \multicolumn{1}{c|}{0.635} & 27.46                           & \multicolumn{1}{c|}{\second 0.828}                & 22.15        & 0.719         \\

\multicolumn{1}{l|}{TiNeuVox}                                     & 24.20 & \multicolumn{1}{c|}{0.616} & 21.28        & 0.307    & \best 22.80 & \multicolumn{1}{c|}{\best 0.725} & 28.22                           & \multicolumn{1}{c|}{0.785}                & 24.50        & 0.646         \\ \hline

\multicolumn{1}{l|}{D3DGS}                                        & 22.40 & \multicolumn{1}{c|}{0.598} & 20.48        & 0.313    & 20.38 & \multicolumn{1}{c|}{0.644} & 22.64                           & \multicolumn{1}{c|}{0.601}                & 26.10        & 0.832        \\

\multicolumn{1}{l|}{4DGaussians}                                  & \second 25.03 & \multicolumn{1}{c|}{\second 0.682} & \best 22.01        & \second 0.367  & 21.99 & \multicolumn{1}{c|}{0.705} & \second 28.47                           & \multicolumn{1}{c|}{0.806}                & \second 27.28        & \second 0.845          \\ 

\multicolumn{1}{l|}{\textbf{Ours}} & \best 25.43 & \multicolumn{1}{c|}{\best 0.697} & \second 21.84 & \best 0.371 & \second 22.34 & \multicolumn{1}{c|}{\second 0.715} & \best 28.75 & \multicolumn{1}{c|}{\best 0.836} & \best 28.80 & \best 0.867 \\                               
\end{tabular}
}
\label{tab:supp_hypernerf}
\end{table}

%%%%%%%%%%%%%%%%%%%%%%%%%%%%%%%%%%%%%%%%%%%%%%%%
%%%%%%%%%%%%%%%%%%%%%%%%%%%%%%%%%%%%%%%%%%%%%%%%

\end{appendix}

\ifx \supp \undefined
    \end{document}
\fi
\fi
\end{document}